\pdfoutput=1
\documentclass[11pt, logo, onecolumn, copyright, colorlinks=true, allcolors=blue]{nvidia_tech_report}

\usepackage{nicefrac}
\usepackage{algorithm}
\usepackage{wrapfig}
\usepackage{algpseudocode}
\usepackage{subcaption}
\title{SOAP, Muon, and Beyond: Pushing LLM Pretraining Scales}

\author{
Mikail Khona, Aditya Vavre, Boxiang Wang, Deyu Fu, Hao Wu, Mike Chrzanowski, Bryan Catanzaro, Dheevatsa Mudigere, Jeff Pool, Michael Lightstone, Mohammad Shoeybi, Mostofa Patwary, Nima Tajbakhsh, Tijmen Blankevoort \\
  \textbf{NVIDIA} \\
\texttt{\{mkhona, avavre, boxiangw, deyuf, skyw, mchrzanowski,}   \texttt{bcatanzaro, dmudigere, jpool, mlightstone, mshoeybi, mpatwary, ntajbakhsh, tblankevoort\}@nvidia.com}
}

\begin{document}

\begin{abstract}
Higher-order optimizers such as Muon and SOAP offer faster convergence than AdamW, but their computational cost and numerical stability challenges have limited adoption at scale. In this work, we adapt and enhance preconditioned gradient methods to overcome the practical challenges of large-scale LLM pretraining.

We first identify instabilities in SOAP at large batch sizes and propose algorithmic modifications including per-step QR orthogonalization and improved preconditioning strategies that eliminate loss spikes and enable stable training in these regimes. We then present a unified empirical study of SOAP, Muon, and AdamW using update-RMS matching to ensure fair learning rate transfer across optimizers. As part of this analysis, we empirically evaluate the orthogonalization quality of Muon. Our experiments on multi-billion-parameter models trained on trillions of tokens reveal that SOAP and Muon consistently outperform AdamW at the scales we tested. Notably, at batch sizes of up to 100M tokens for next-token prediction, these optimizers maintain training stability and quality while AdamW degrades.

To enable efficient training at large scale, we introduce a layer-wise distributed optimizer compatible with Megatron-LM. Our implementation balances memory and hides communication while avoiding approximations to the optimizer computations, thus retaining their convergence benefits. Additionally, we identify and build specific system-level improvements to further accelerate our layer-wise implementation. To support the research community, we release a codebase that contains emerging algorithms for optimization:

\texttt{https://github.com/NVIDIA-NeMo/Emerging-Optimizers}
\end{abstract}

\maketitle

\section{Introduction}

The choice of the optimization algorithm remains at the center of frontier model training frameworks. Indeed, it serves as the heart of all large-scale model training, linking systems engineering with algorithms and model convergence. From a systems perspective, the optimizer dictates the fundamental constraints of distributed execution: optimizer states typically consume significantly more memory than the model parameters themselves, heavily influencing model sharding and memory management strategies. Further, an optimizer's ability to maintain stability at extreme batch sizes dictates how far training can scale across large accelerator clusters without being bottlenecked by communication overhead due to the synchronous nature of training workloads. Algorithmically, the optimizer governs data efficiency, convergence rates, and the generalization ability of the model.

Despite this dual impact, the evolution of optimization in deep learning has been defined by a practical tension: the computational simplicity and scalability of element-wise scalar optimizers, such as AdamW \cite{kingma2014adam, loshchilov2017decoupled}, RMSProp~\cite{hinton2012neural}, and LaProp~\cite{ziyin2020laprop}, have largely overshadowed curvature-aware higher-order methods.
While first-order methods like AdamW dominate due to their ease of use, they inherently treat individual parameter element updates independently, ignoring the correlational structure between gradients and the operator nature of the corresponding neural network weights. Conversely, higher-order optimizers (approximating second-order information) capture the geometry of the loss landscape \cite{george2018fast, martens2015optimizing, bollapragada2018progressive}, theoretically allowing for larger steps, faster convergence, and potentially higher accuracy, but suffer from a lack of scalability due to their complexity. 

Recently, tensor-wise adaptive methods such as the Shampoo optimizer and modern variants like SOAP \cite{vyas2024soap}, Eigen-corrected Shampoo \cite{eschenhagen2025purifying}, and KL-Shampoo \cite{lin2025understanding}, along with spectral optimizers such as Muon and Scion \cite{pethick2025training}, have emerged as a middle ground, offering the benefits of second-order methods with manageable computational and memory overhead. However, these methods face scalability hurdles when applied to large frontier models like fine-grained MoEs.

The rest of this paper is organized around the formulation, empirical evaluation, and systems engineering required to scale these higher-order optimizers. Our core contributions are as follows:
\begin{itemize}
    \item \textbf{Large Batch-Size Scaling for MoEs with Muon}: We use an update-RMS matching framework to fairly transfer learning rates and compare AdamW, Muon, and SOAP optimizers. We motivate the MoE large-batch setting in Sec.~\ref{sec:large-batch-pretraining} and present empirical results in Sec.~\ref{sec:muon-large-batch} and Sec.~\ref{sec:compare-muon-soap}. Our evaluation on up to 72-billion parameter MoE models demonstrates that while AdamW's performance degrades beyond critical batch sizes, both Muon and SOAP maintain token-efficient convergence at global batch sizes up to 100M tokens.
    \item \textbf{Fixing Instability for SOAP's Preconditioner Computation}: We identify a critical instability in the standard implementation of SOAP \footnote{\url{https://github.com/nikhilvyas/SOAP}} unique to the large-batch regime, caused by a lag between the preconditioner and current gradient statistics. Sec.~\ref{sec:soap-stability} shows that enforcing the per-step eigenbasis updates with current gradients and integrating a recently developed KL-divergence-based covariance estimation method for accumulating SOAP's Kronecker factors~\cite{lin2025understanding} eliminates all training loss spikes and improves training stability, allowing for more confidence for scaling to larger batch and model sizes in production.
    \item \textbf{Comparing SOAP and Muon}: We compare the accuracy and stability of these optimizers in the large batch setting, finding that both Muon and SOAP outperform AdamW and KL-SOAP has a slight edge over Muon. We also discuss the limitations of our conclusions. 
    \item \textbf{Scalable Parallelization}: We present an efficient, layer-wise distributed implementation designed to enable scalability within the Megatron-LM framework. We briefly discuss the distributed optimizer constraints in Sec.~\ref{sec:large-scale-preconditioning} and present the implementation in Sec.~\ref{sec:layerwise-optimizer}.
    \item \textbf{\texttt{Emerging-Optimizers}}: We contribute an open-source library for optimizer research that includes implementations of all optimizers considered in this report as well as many experimental variants. 
\end{itemize}

\section{A Primer on Adam, Muon and SOAP}

Adam-style optimizers are the most common optimizers used in modern LLM pretraining, which extend stochastic gradient descent (SGD) with coordinate-wise adaptive learning rates and decoupled weight decay~\cite{kingma2014adam, loshchilov2017decoupled}. Here we provide a very brief overview of these methods and how they are related.

Let $g_t \in \mathbb{R}^{d=mn}$ be the flattened gradient vector $G_t \in \mathbb{R}^{m\times n}$ of a parameter tensor at step $t$. AdamW maintains exponential moving averages (EMA) of the gradient and squared gradient for each parameter element independently with $\beta_1, \beta_2$ being the EMA timescale coefficients.
\begin{equation*}\label{adam_momentum}
    m_t = \beta_1 m_{t-1} + (1 - \beta_1) g_t \\
    v_t = \beta_2 v_{t-1} + (1 - \beta_2) g_t \odot g_t 
\end{equation*}
where $\odot$ denotes the element-wise (Hadamard) product. The preconditioned update direction for AdamW (ignoring first-moment momentum $m_t$ and bias corrections for simplicity) applies a diagonal preconditioner to the gradient:
\begin{equation*}
    u_t = m_t \circ \text{diag}\left( \dfrac{1}{(v_t)^{1/2} + \epsilon} \right)
\end{equation*}
This element-wise structure makes AdamW memory-regular, and easy to shard in large distributed systems.
However, it also means that AdamW ignores the matrix or tensor structure of neural network weights: each coordinate is rescaled separately, but correlations across rows, columns, heads, experts, or projection subspaces are not modeled explicitly.

A separate line of optimizers attempts to exploit this structure through preconditioning.
Shampoo~\cite{gupta2018shampoo} approximates second-order information by maintaining Kronecker-factored gradient statistics for each tensor dimension, allowing matrix- or tensor-shaped parameters to be updated using curvature-aware transformations without forming a full dense preconditioner. Shampoo maintains two separate symmetric covariance matrices for the rows and columns:
\begin{align*}
    L_t &= \beta_2 L_{t-1} + (1 - \beta_2) G_t G_t^\top \\
    R_t &= \beta_2 R_{t-1} + (1 - \beta_2) G_t^\top G_t
\end{align*}
Shampoo's update direction is then produced through a non-diagonal kronecker-factored preconditioner, derived from inverse powers of the kronecker factors:
\begin{equation*}
    u_t = (R_t^{-1/4} \otimes L_t^{-1/4}) g_t
\end{equation*}

SOAP (Alg.\ref{soap_algo}) builds on Shampoo by combining Shampoo-style preconditioning with Adam-like elementwise adaptive updates in the preconditioner eigenbasis ~\cite{vyas2024soap}:
\begin{equation}
    u_t = Q_L\texttt{Adam}(Q_L^Tm_tQ_R)Q_R^T
\end{equation}

Intuitively, SOAP first rotates gradients into a basis where their row and column correlations are approximately diagonalized (i.e. the eigenbasis of $R_t$ and $L_t$), applies an adaptive element-wise update in that basis, and then rotates the update back out of the preconditioner eigenbasis.
This gives SOAP a closer connection to AdamW than classical Shampoo, since for identity-valued eigenvectors, we exactly recover AdamW, while retaining the ability to exploit structure within weight matrices, possibly allowing for smoother hyperparameter transfer. However both Shampoo and SOAP maintain full precision preconditioner and kronecker factor matrices requiring significantly more memory than AdamW. 

Muon (Alg.\ref{muon_algo}) skips computing and storing preconditioners and follows a different strategy. Rather than estimating preconditioners and full adaptive second-moment statistics in a preconditioned basis, Muon applies momentum like AdamW and Shampoo, Eq.\ref{adam_momentum}, and then orthogonalizes it.
To orthogonalize $M_t$ and approximate its polar factor (the closest orthogonal matrix, equivalent to $U V^\top$ from the singular value decomposition $M_t = U \Sigma V^\top$), Muon employs Newton-Schulz iterations~\cite{pethick2025training, liu2025muon, shah2025practical, amsel2025polar, jordanmuon} which use matrix polynomial iteration to approximately compute the polar factor.

This can be viewed as a spectral update, since for 2D weight matrices, Muon changes the direction of the update so that its singular values are controlled, rather than applying independent coordinate-wise rescaling like AdamW.
In simplified terms, SOAP/Shampoo-style whitening and Muon-style orthogonalization are closely related: both reduce the dominance of large singular directions and produce updates that better respect the operator structure of linear layers. Notably, turning off EMA by setting $\beta_1, \beta_2 = 0$ in Shampoo mathematically reduces its preconditioned update directly to the polar factor $U V^\top$, since $(\mathbf{GG^T})^{-1/4} \, \mathbf{G} \, (\mathbf{G^T G})^{-1/4} = U V^\top$, recovering the Muon update. See Appendix \ref{svd_muon_soap} for a detailed SVD interpretation of SOAP's preconditioning and the intuitive relationships between all three optimizers.

There are practical differences between using these optimizers at scale. Here we summarize the qualitative differences.
AdamW is the easiest optimizer to scale because its state and computation are element-wise and hence infinitely shardable.
SOAP is more expressive, but it must maintain and update Kronecker factors and eigenbases, making it more sensitive to numerical stability, preconditioner freshness, has a much larger memory footprint, and complicated matrix-based distributed tensor layout for sharding.
Muon has a lower footprint than AdamW since it does not maintain second moments but it still requires full 2D matrix updates which faces the same matrix-based sharding issues as SOAP. Muon uses more computation in the optimizer step and this overhead depends on the quality and cost of orthogonalization which scales with number of Newton-Schulz iterations. 

\section{A Primer on Optimization for Large Scale, Large Batch Pretraining}\label{sec:large-batch-pretraining}

\subsection{Batch Size Scaling for Mixture-of-Experts Models}\label{sec:moe-batch-scaling}
Large-batch scaling is particularly nuanced for Mixture-of-Experts (MoE) models because the majority of the model's parameters reside in the expert linear layers (often $>90\%$ of the total parameter count). Unlike dense models, these layers observe a significantly lower \textit{effective} batch size due to sparse top-k routing. Consider a global batch size of $B_{\text{Global}}$ tokens trained with a top-$k$ routing strategy over $N$ total experts. Assuming an ideally load-balanced router, the effective batch size $B^{\text{expert}}_{\text{eff}}$ seen by any single expert is given by:
\begin{equation}B^{\text{expert}}_{\text{eff}} = B_{\text{Global}} \times \frac{k}{N}\end{equation}
Since $k \ll N$ in typical fine-grained sparse MoE configurations (e.g., top-8 gating with 256 experts \cite{liu2024deepseek}, leading to ${B^{\text{expert}}_{\text{eff}}}/{B_{\text{Global}}}= 0.03125$), each expert sees only a small fraction of the global batch. Consequently, when scaling to very large global batch sizes, the sparse expert parameters remain in a lower-effective-batch regime that is generally easier to optimize. In contrast, the dense and shared parameters must process the full global batch. As a result, increasing the global batch size primarily stresses the large-batch tolerance of the dense components. This makes MoE training highly reliant on optimization methods that can maintain stability and token efficiency specifically for these dense parameters operating at extreme batch sizes \cite{sun2024hunyuan}.

\subsection{Batch Size--Learning Rate Scaling and Update Variance}\label{sec:batch-lr-scaling}

Because our comparisons change global batch size by up to 4x, we need a learning-rate transfer rule that keeps update scale comparable before attributing differences to the optimizer itself. In the context of SGD, modifying the mini-batch size requires a corresponding adjustment to the learning rate to maintain stable and transferrable optimization dynamics. Let us consider a simple SGD update where the gradient estimator $\hat{g}_B$, computed over a mini-batch of size $B$, is given by:
$$\hat{g}_B = \frac{1}{B} \sum_{i=1}^B \nabla L_i(\theta)$$

The parameter update is defined as $\Delta \theta = \eta \hat{g}_B$, where $\eta$ represents the learning rate. Assuming the individual sample gradients are independent, the covariance of the gradient estimator scales inversely with the batch size:
$$\mathrm{Cov}(\hat{g}_B) = \frac{1}{B} \Sigma$$
where $\Sigma$ is the covariance matrix of the per-sample gradients. Consequently, the covariance of the parameter update itself is:
$$\mathrm{Cov}(\Delta \theta) = \eta^2 \frac{1}{B} \Sigma$$

When scaling the batch size from an initial size $B$ to a target size $B'$, a heuristic approach is to preserve the variance scale of the parameter updates. We equate the covariances of the updates under both batch size regimes:
$$\frac{\eta^2}{B} \Sigma = \frac{(\eta')^2}{B'} \Sigma$$

Solving for the updated learning rate $\eta'$ yields the Square Root Scaling Rule:
\begin{equation}
\eta' = \eta \sqrt{\frac{B'}{B}}
\end{equation}

To ensure that the scale of the random fluctuations in parameter updates ($\Delta \theta$) remains similar when we switch from batch size $B$ to $B'$, we adjust the learning rate by the square root of the ratio of the batch sizes. The logic behind this rule is focused on noise or variance preservation. Although the linear scaling rule \cite{goyal2017accurate} has also been proposed, square-root scaling is safer when making very large leaps in batch size, as it prevents the learning rate from exploding. Several more rigorous treatments of this rule have been developed, such as in \cite{malladi2022sdes}.

\subsection{Strategies for Enabling Distributed Training with Preconditioning}\label{sec:large-scale-preconditioning}

The gains from large batch training properties of Muon and SOAP can only be realized if the training system can preserve full matrix structure under tensor parallelism without additional exposed communication. When tensor parallelism is applied, the weights of a linear layer are distributed across multiple GPUs. Beyond the simplest method of preconditioning based solely on the weights owned by a given GPU, commonly referred to as ``blocking'', \texttt{Emerging-Optimizers} also supports two distinct approaches for preconditioning on the entire layer's weights:
\begin{itemize}
    \item Duplicated: The weights are initially all-gathered to all GPUs within the Tensor Parallelism (TP) domain. Subsequently, each GPU independently executes the Newton-Schulz iteration. This mode is preferable for smaller layers where communication is the limiting factor.
    \item Distributed: In each Newton-Schulz iteration, the intermediate results of the first matrix multiplication are all-reduced. This mode is preferable for larger layers where computation is the limiting factor.
\end{itemize}

In both the ``duplicated'' and ``distributed'' modes, the inputs are first normalized using the statistics of the entire layer's weight, thereby ensuring mathematical equivalence to the scenario without tensor parallelism.

Conventional distributed optimizer sharding approaches (like ZeRO-1 \cite{zhao2023pytorch, rajbhandari2020zero}) are
not directly appropriate for matrix-based optimizers like Muon and SOAP. If weights and optimizer states are evenly distributed among DP ranks, updates cannot be calculated based on the shard available on each rank. Additional communication is needed to collect tensors for calculating the full update. To overcome this, we developed a layer-wise distributed optimizer, in which parameters of different layers are distributed to different DP ranks. Each GPU has full layers' worth of parameters so that the preconditioner can be calculated\footnote{\url{https://docs.nvidia.com/nemo/emerging-optimizers/latest/primer/layerwise-distributed-optimizer.html}}.

\section{Background and Related Work}\label{sec:background}
\subsection{Shampoo and related optimizers}
Shampoo \cite{gupta2018shampoo} is an optimization algorithm that bridges the gap between first-order methods like Adam and full-matrix methods like Natural Gradient Descent and Adagrad. At its core, Shampoo maintains a running estimate of the all-dimensional correlations of the parameter gradients, but it does so efficiently by using a Kronecker-factored approximation of the Fisher information or the empirical covariance matrix. By treating gradients as tensors rather than flattened vectors, it preserves the spatial or structural correlations inherent in neural network layers.
Building upon this foundational work, \cite{anil2020scalable} develops techniques to scale Shampoo to larger models, including ``blocking" or restricting preconditioning to subsets of matrices and coupled iterations \cite{anil2020scalable, modoranu2026dash} for root-inverse computation that avoid eigensolvers. These optimizations effectively mitigate the memory and computational bottlenecks traditionally associated with Kronecker-factored second-order methods. By amortizing the cost of the preconditioner update across multiple steps and leveraging hardware-accelerated matrix operations, these variants achieve a convergence rate superior to first-order optimizers like Adam, while maintaining a comparable per-step wall-clock time. 
Modern variants of Shampoo have been proposed, such as Eigen-decomposed Shampoo \cite{eschenhagen2025purifying}, KL-Shampoo \cite{lin2025understanding}, SOAP \cite{vyas2024soap}, and DASH \cite{modoranu2026dash}. These algorithms improve the estimation of Shampoo's preconditioner and thus improve convergence and stability. 

Spectral descent was first developed in \cite{carlson2015preconditioned, carlson2015stochastica, carlson2015stochasticb, flynn2017duality}, and a rigorous framework for spectral descent for all layer types in deep learning was established in \cite{pethick2025training}. In another line of work, Bernstein \textit{et al}. \cite{bernstein2024modular, bernstein2024old, large2024scalable} have shown that steepest descent under different norm constraints on parameters provides a picture that unifies element-wise optimizers such as AdamW and matrix-based optimizers such as Muon. Several variations of Muon have recently been developed, such as Muon+ \cite{zhang2026muon+}, Muon++ \cite{zhao2026towards}, AdaMuon~\cite{si2025adamuon}, NorMuon~\cite{li2025normuon}, Muon-Hyperball~\cite{wen2025hyperball_part1}, Muon on the spectral sphere \cite{xie2026controlled}, NAMO~\cite{zhang2026adam}, SPEL \cite{yang2026manifold}, Newton-Muon \cite{du2026newton}, and Mousse \cite{zhang2026mousse}.

\subsection{Systems to enable higher-order optimizers}\label{sec:related-systems}
Conventional Fully Sharded Data Parallel (FSDP) systems restrict tensors to symmetric element-wise shards. While this is highly effective for standard element-wise optimizers like AdamW, it fundamentally conflicts with higher-order, structure-aware optimizers. These optimizers often require access to full 2D parameter matrices (e.g., for computing Muon's Newton-Schulz matrix-sign polynomial iteration) or specific contiguous blocks. 
To address this, veScale-FSDP \cite{wang2026vescale} introduces \texttt{RaggedShard}, a flexible sharding format that supports asymmetric and arbitrary data granularities. This enables an execution pipeline where full parameters can be dynamically gathered to a root device for complex local computations, and the resulting updates asynchronously scattered back. To support this flexibility without sacrificing scale, veScale-FSDP pairs \texttt{RaggedShard} with a structure-aware planning algorithm and a zero-copy distributed buffer, minimizing both communication overhead and memory fragmentation.
Canzona \cite{wang2026canzona} builds an asynchronous framework that decouples the logical assignment of optimizer states from the physical distribution of model parameters. For Data Parallelism (DP), it employs a static partitioning strategy that maintains matrix atomicity for the optimizer while preventing load imbalance. For Tensor Parallelism (TP), Canzona utilizes an asynchronous pipeline with micro-group scheduling to batch fragmented updates and hide matrix reconstruction overheads. This approach masks the communication costs of higher-order optimizers, reducing the latency of each step while preserving the throughput of established parallel architectures.

\section{Pretraining Experiments with Muon and SOAP}\label{sec:experiments}

Next, we describe our setup and follow up with experiments with Muon, SOAP and experiments comparing Muon and SOAP in the large batch regime on dense and MoE models. 

\subsection{Model and Dataset}

\paragraph{Model Architecture} We run all of our optimizer experiments on 8B dense GPT models \cite{brown2020language} and 3B active/30B transformer-only MoEs \cite{yang2025qwen3}, and finally scale up to 8B active/72B hybrid Mamba-Transformer MoEs \cite{blakeman2025nemotron}.

\paragraph{Dataset} We use 1T- and 3T-token subsets of the full Nemotron-3 dataset \cite{blakeman2025nemotron}.

\paragraph{Architecture Configurations} 
The core architectural hyperparameters for our dense and sparse models are detailed in Table~\ref{tab:core_architecture}. For the attention mechanism, all models utilize a head dimension of 128 and Grouped-Query Attention (GQA) to improve inference efficiency. 

\begin{table}[htbp]
\centering
\makebox[\textwidth][c]{
    \begin{tabular}{lcccccc}
    \toprule
    \textbf{Model} & \textbf{Hidden Size} & \textbf{Layers} & \textbf{GQA} & \textbf{FFN / Expert Size} \\
    \midrule
    Nemotron-3-Nano-30B-A3B & 2688 & 52 & 2 & 1856 \\
    8B Dense Transformer & 4096 & 32 & 8 & 21504 \\
    Nemotron-3-72B-A8B & 4096 & 52 & 8 & 2688 \\
    Qwen3-30B-A3B & 2048 & 48 & 4 & 768 \\
    \bottomrule
    \end{tabular}
}
\caption{Core architectural hyperparameters for the evaluated models. For MoE models, the FFN size denotes the hidden size of an individual expert. All models use 32 heads with head dimension 128.}
\label{tab:core_architecture}
\end{table}

\begin{table}[htbp]
\centering
\makebox[\textwidth][c]{
    \begin{tabular}{lcccc}
    \toprule
    \textbf{Model} & \textbf{Total Experts} & \textbf{Active Experts (Top-$k$)} & \textbf{Shared Expert Size} \\
    \midrule
    Nemotron-3-Nano-30B-A3B & 128 & 6 & 3712 \\
    Nemotron-3-72B-A8B  & 512 & 6 & 5376 \\
    Qwen3-30B-A3B  & 128 & 8 & N/A \\
    \bottomrule
    \end{tabular}
}
\caption{MoE and routing configurations for the Mixture of Experts (MoE) models.}
\label{tab:moe_config}
\end{table}

\paragraph{Mixture of Experts (MoE) Configuration}
For our sparse models, we employ a load-balancing strategy utilizing a sigmoid score function combined with sequence-level auxiliary loss (\texttt{seq\_aux\_loss}). By applying an auxiliary loss coefficient of $1 \times 10^{-4}$, we effectively mitigate routing collapse—penalizing unbalanced token assignments and ensuring that active experts receive a relatively uniform distribution of tokens across the sequence. Our MoE variants also leverage shared experts with overlapping computation. For instance, the 8B Hybrid MoE scales up to 512 total experts with a top-6 routing sparsity granularity and an additional shared expert intermediate size of 5376.

\begin{table}[h]
\centering
\begin{tabular}{lcc}
\toprule
\textbf{Model} & \textbf{Max LR} & \textbf{Min LR} \\
\midrule
Nemotron-3-Nano-30B-A3B & $1.0 \times 10^{-3}$ & $1.0 \times 10^{-5}$ \\
8B Dense GPT & $8.0 \times 10^{-4}$ & $8.0 \times 10^{-6}$ \\
Nemotron-3-72B-A8B & $8.0 \times 10^{-4}$ & $8.0 \times 10^{-6}$ \\
Qwen3-30B-A3B & $1.6 \times 10^{-3}$ & $1.6 \times 10^{-5}$ \\
\bottomrule
\end{tabular}
\caption{Learning rate schedules for the evaluated models. All models utilize a global batch size of 3072 (micro-batch size of 1), a constant weight decay of 0.1, and a Warmup-Stable-Decay (WSD) schedule with a minus-square-root (\texttt{minus\_sqrt}) decay style.}
\label{tab:lr_schedules}
\end{table}

\paragraph{Optimization and Learning Rate Schedule}
To train our models, the baselines use a standardized global batch size of 3072 samples, with a sequence length of 8192 tokens each (leading to a global batch size of 25M tokens) across all architectures. We apply a uniform weight decay of 0.1. For learning rate progression, we adopt a Warmup-Stable-Decay (WSD) scheduling strategy \cite{hagele2024scaling}. Following a linear warmup phase, the learning rate is held constant during the stable phase before transitioning into an annealing phase. Specifically, we utilize a minus-square-root (\texttt{minus\_sqrt}) decay style for the final WSD decay.

\subsection{Optimizer Comparison Protocol}\label{sec:optimizer-comparison-protocol}

A critical challenge in comparing optimizers is ensuring a fair hyperparameter configuration. To address this, we utilize an \textbf{update RMS matching} framework pioneered by Kimi-Moonshot \cite{liu2025muon}. This technique allows us to transfer learning rates (LRs) effectively between optimizers by ensuring the root mean square (RMS) norm of the parameter updates is consistent across methods, reducing tuning compute.

Under this RMS-matching framework, SOAP and AdamW naturally have identical update RMS norm since rotation matrices do not change norm:
\begin{align*}
\|\mathbf{\Delta W}\|_F^2 &= \text{Tr}\left( \mathbf{\Delta W}^T \mathbf{\Delta W} \right) \\
&= \text{Tr}\left( (\mathbf{Q}_L \mathbf{N} \mathbf{Q}_R^T)^T (\mathbf{Q}_L \mathbf{N} \mathbf{Q}_R^T) \right) \\
&= \text{Tr}\left( (\mathbf{Q}_R^T \mathbf{Q}_R) \mathbf{N}^T (\mathbf{Q}_L^T \mathbf{Q}_L) \mathbf{N} \right) \\
\text{Since $\mathbf{Q}_L$ and $\mathbf{Q}_R$ are orthonormal, }  \\\mathbf{Q}_L^T \mathbf{Q}_L &= \mathbf{I}, \mathbf{Q}_R^T \mathbf{Q}_R = \mathbf{I}, \\
\|\mathbf{\Delta W}\|_F^2 &= \|\mathbf{N}\|_F^2
\end{align*}

We explicitly apply Kimi’s update RMS matching framework to Muon (Algorithm~\ref{algorithm_muon}), which gives us a correction factor that depends on the EMA damping factor of momentum \cite{kexuefm-11307, li2024stochastic, kosson2023rotational}:
\begin{equation*}
    \sqrt{\frac{1-\beta_1}{1+\beta_1}} \approx 0.2
\end{equation*}
Notably, we observed that using Nesterov momentum \cite{sutskever2013importance, dozat2016incorporating} did not improve convergence and skipped using it for both Muon and SOAP, and using simple exponential moving averages (EMA) for momentum. 

\subsection{Muon Scales Better than AdamW at Large Batch Sizes for Pretraining}\label{sec:muon-large-batch}
Prior research indicates that Muon and its variants significantly outperform AdamW as the global batch size scales \cite{pethick2025training}. While AdamW often hits a "critical batch size" \cite{mccandlish2018empirical} beyond which increasing parallelism yields diminishing returns, token inefficiency, or instability, Muon's orthogonalization and spectral descent allow it to maintain convergence at much larger batch sizes.

\begin{figure}[htbp]
    \centering
    % First figure
    \begin{minipage}{\textwidth}
        \centering
        \includegraphics[width=0.8\linewidth]{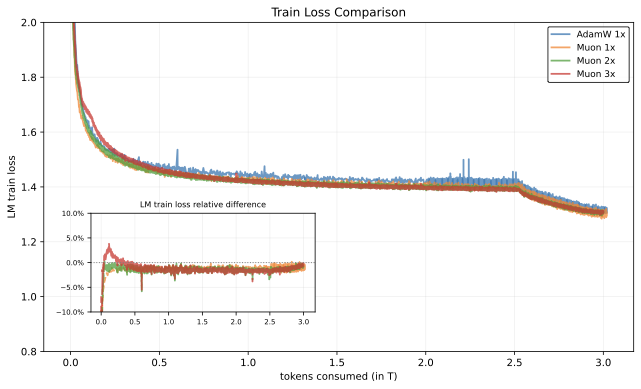}
        \caption{Loss curves of Muon at large batches (2x and 3x) with NVIDIA-Nemotron-Nano-V3-30B-A3B on a 3T-token horizon compared against the original AdamW baseline of \cite{blakeman2025nemotron}. The inset shows the relative loss gaps as a function of tokens against the AdamW baseline. Larger batches typically underperform and eventually exceed lower batch sizes, especially as the token horizon scales.}
        \label{fig:nanov3-baseline}
    \end{minipage}
    \vspace{2ex} % Vertical space between the figures

    % Second figure
    \begin{minipage}{\textwidth}
        \centering
        \includegraphics[width=0.8\linewidth]{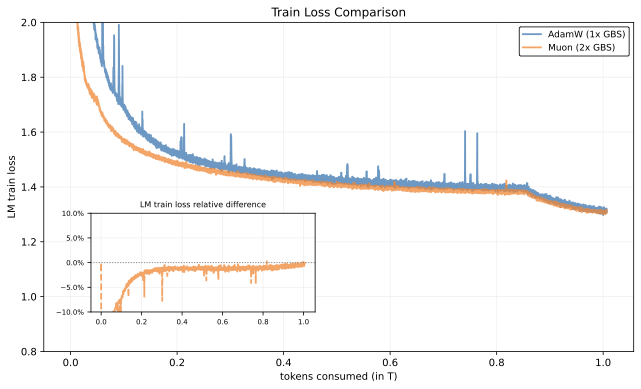}
        \caption{Hybrid Mamba-MoE 72B-A8B trained on 1T tokens with Muon at 2x the batch size of the AdamW baseline (25M tokens). Inset shows that Muon at larger batches has \textit{lower} loss and no loss spikes relative to the smaller batch AdamW baseline. }
        \label{fig:latent-8b-lr}
    \end{minipage}
\end{figure}

Our experiments confirm that Muon provides better stability and lower loss compared to AdamW in this setting. At a baseline ($1\times$) batch size, Muon achieves lower training loss and exhibits fewer spikes, Fig.\ref{fig:nanov3-baseline}. We have empirically found training loss to correlate to held-out validation loss on our dataset and we include evaluation on downstream tasks for a subset of models in Table \ref{tab:gbs_ablation_results}. Crucially, as we scale the global batch size, AdamW fails to scale effectively, rapidly encountering instabilities and diminishing returns. In contrast, Muon maintains robust convergence characteristics, effectively utilizing the increased data parallelism to speed up training. 

\begin{minipage}{\linewidth}
\begin{algorithm}[H]
\caption{Muon: \textbf{M}oment\textbf{U}m \textbf{O}rthogonalized by \textbf{N}ewton-Schulz with Weight Decay}
\label{muon_algo}
\label{algorithm_muon}
\begin{algorithmic}[1]
\State \textbf{Hyperparameters:} $\eta_t,\, \beta_1,\, \lambda_t,\, n,\, \{(a_i,b_i,c_i)\}_{i=1}^{n},\, \epsilon$
\State \textbf{Optimizer States:} 
$
    \mathbf{M}_t \in \mathbb{R}^{\text{in}\times \text{out}} \text{ (Momentum buffer)}
$
\State $\mathbf{G}_t = \nabla_{\mathbf{W}} L(\mathbf{W}_{t-1})$ 

\texttt{Update Momentum Buffer with EMA:}
\State $\mathbf{M}_t \gets \beta_1 \mathbf{M}_{t-1} + (1-\beta_1)\mathbf{G}_t$ 

\texttt{Newton-Schulz Orthogonalization:}
\State $\mathbf{X}_0 \gets \mathbf{M}_t / \max(\|\mathbf{M}_t\|_F, \epsilon)$ \Comment{Initialize with normalized momentum}
\For{$i = 1$ \textbf{to} $n$}
    \State $\mathbf{A} \gets \mathbf{X}_{i-1} \mathbf{X}_{i-1}^T$
    \State $\mathbf{B} \gets b_i \mathbf{A} + c_i \mathbf{A}^2$
    \State $\mathbf{X}_i \gets a_i \mathbf{X}_{i-1} + \mathbf{B} \mathbf{X}_{i-1}$
\EndFor
\State $\mathbf{O}_t \gets \mathbf{X}_n$ \Comment{Resulting orthogonalized update direction}

\texttt{Post-processing update direction normalization:}
\State $\mathbf{\Delta W} \gets \mathbf{O}_t \cdot \sqrt{\max(\text{in}, \text{out})}$ \Comment{Kimi scaling for update RMS matching}

\texttt{Decoupled weight decay and parameter update:}
\State $\mathbf{W}_{t} \gets (1 - \eta_t \lambda_t) \mathbf{W}_{t-1} - \eta_t \mathbf{\Delta W}$

\end{algorithmic}
\end{algorithm}
\end{minipage}

\begin{table}[h]
\centering
\small
\begin{tabular}{lll}
\toprule
\textbf{Hyperparameter} & \textbf{Description} & \textbf{Value} \\
\midrule
$\beta_1$ & Momentum EMA coefficient & 0.9 \\
$\lambda_t$ & Decoupled weight decay coefficient & 0.1 \\
$n$ & Number of Newton-Schulz iterations & 16 \\
$\{(a_i,b_i,c_i)\}_{i=1}^{n}$ & Newton-Schulz polynomial coefficients & PolarExpress \cite{amsel2025polar} \\
$\epsilon$ & Lower bound for the Frobenius-norm normalization & $10^{-7}$ \\
\bottomrule
\end{tabular}
\caption{Muon hyperparameters used in Algorithm~\ref{algorithm_muon}.}
\label{tab:muon_hparams}
\end{table}

We hypothesize that at extreme batch sizes, the gradient noise decreases, which can cause AdamW's element-wise second-moment estimator ($\mathbf{v}_t$) to become poorly calibrated or overly confident, leading to suboptimal step sizes. In contrast to AdamW, Muon's spectral preconditioning (i.e. orthogonalization of the momentum matrix) structurally normalizes the update directions \textit{independent} of coordinate-wise scaling, making it resilient to the variance fluctuations of large-batch scaling.

\begin{figure}[htbp]
    \centering
    \begin{subfigure}[b]{0.48\linewidth}
        \includegraphics[width=\linewidth, keepaspectratio]{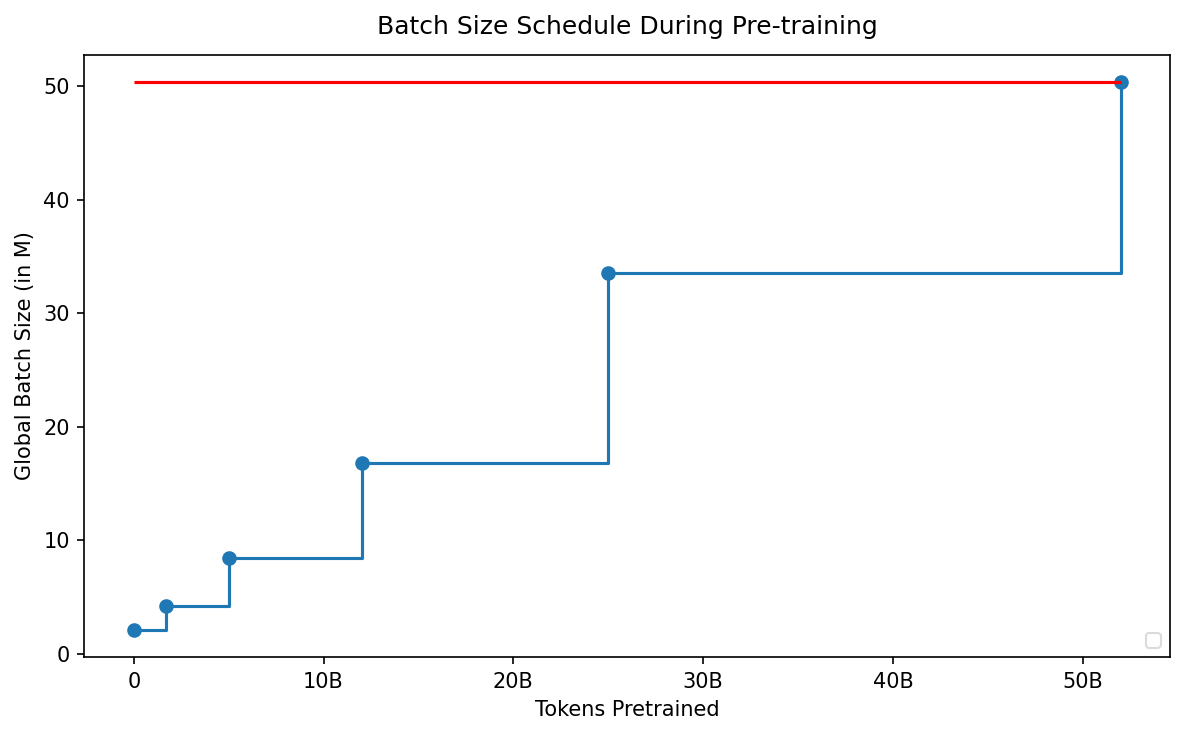}
        \label{fig:batch_schedule_50M}
    \end{subfigure}
    \hfill
    \begin{subfigure}[b]{0.48\linewidth}
        \includegraphics[width=\linewidth, keepaspectratio]{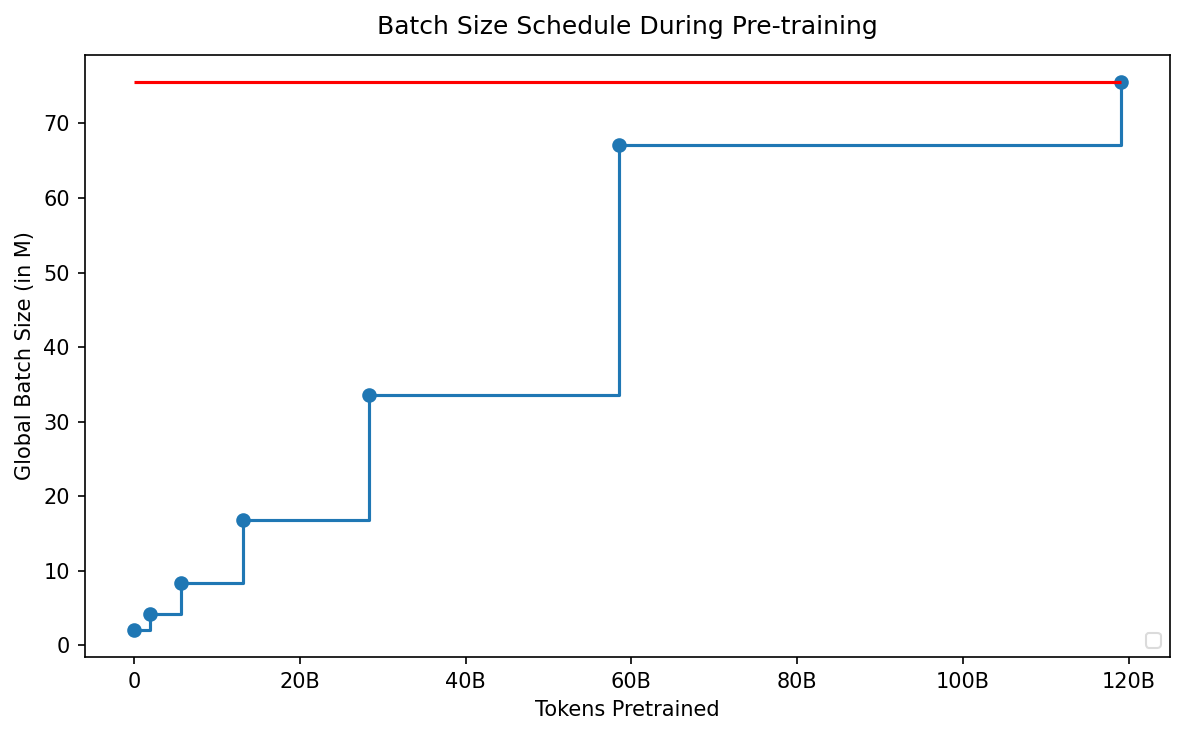}
        \label{fig:batch_schedule_75M}
    \end{subfigure}
    \caption{The batch size ramp-up schedules (in blue) used in pretraining experiments for final batch sizes of 50M and 75M tokens (in red). }
    \label{fig:batch_schedules}
\end{figure}

We also find in Fig.~\ref{fig:latent-8b-lr} that implementing a \textbf{batch size ramp-up} phase early in training allows Muon to significantly outperform AdamW on both final training loss and downstream evaluation metrics, while using a larger batch size for the majority of training. We note that the rampup schedule was used only with the larger batch pretraining experiments for both optimizers. Following \cite{li2025minimax}, we adopt a ramp-up schedule governed by a few simple rules. We treat the language modeling loss as the signal dictating the optimal batch size. Starting from a batch size of 256 samples (2M tokens), we double every $S$ steps over a fixed ramp-up horizon of $N$ total steps, irrespective of the token horizon, until the target batch size is reached, Fig.~\ref{fig:batch_schedules}. The learning rate warmup is tied directly to the batch size ramp-up to approximately maintain the square-root scaling ratio throughout, i.e., $\eta \propto \sqrt{B}$, as motivated by large-batch training theory \cite{You2020Large, hoffer2017train, you2017large, krizhevsky2014one}. Specifically, the learning rate at each stage is tuned to preserve the variance scale of the parameter update as the batch size changes. We note that we did not tune the batch size ramp-up schedule as a function of model scale \cite{li2025minimax, mccandlish2018empirical}, where the critical batch size typically depends on the training loss; hence, larger models are expected to reach higher critical batch sizes earlier than smaller models.

We evaluate on two architectures: NVIDIA-Nemotron-Nano-V3, a hybrid attention-mamba MoE model with 30B total and 3B active parameters trained on 3T tokens, and a hybrid LatentMoE model with Multi-Token Prediction (MTP) layers \cite{gloeckle2024better, liu2024deepseek} with 72B total and 8B active parameters trained on a 1T token horizon. Table~\ref{tab:gbs_ablation_results} reports downstream evaluation results for these models. For both architectures, Muon consistently matches or outperforms AdamW at baseline batch size and continues to improve as batch size scales, whereas AdamW does not (Fig.~\ref{fig:sup_adamW_2xgbs}). The gains are most pronounced in coding and commonsense reasoning.
We also note that while ablating the hybrid Mamba-MoE architecture of Table~\ref{tab:gbs_ablation_results}, we found an important architectural sensitivity: Muon is best applied selectively to dense linear projections, as we found that falling back to AdamW for Mamba's \texttt{conv1D} weights \cite{dao2024transformers} yields better convergence.
These results validate that Muon's large-batch efficiency transfers to MoE architectures up to the $\approx5e^{22}$ pretraining FLOPs scale.

We further validate that these results hold with MXFP8 pretraining in Appendix Sec.~\ref{app_sec:mxfp8}. 

 \begin{table}[h]
\centering
\resizebox{\textwidth}{!}{%
\begin{tabular}{lcccccc}
\toprule
\textbf{Metric} 
& \multicolumn{4}{c}{\textbf{NVIDIA-Nemotron-Nano-V3}} 
& \multicolumn{2}{c}{\textbf{8B Hybrid MTP}} \\
\cmidrule(lr){2-5} \cmidrule(lr){6-7}
& Muon 3$\times$ & Muon 2$\times$ & Muon 1$\times$ & AdamW 1$\times$ 
& Muon 2$\times$ & AdamW 1$\times$ \\
\midrule

\multicolumn{7}{c}{\textbf{MMLU}} \\
\midrule
MMLU & 74.00 & \textbf{74.80} & 73.71 & 73.38 & \textbf{74.89} & 74.59 \\
MMLU PRO CoT & 55.61 & 56.77 & \textbf{58.19} & 54.45 & \textbf{59.60} & 57.52 \\

\midrule
\multicolumn{7}{c}{\textbf{Coding}} \\
\midrule
HumanEval & 62.22 & \textbf{64.79} & 62.56 & 63.26 & \textbf{65.79} & 57.62 \\
HumanEval+ & 59.18 & \textbf{62.13} & 59.63 & 58.63 & \textbf{61.86} & 56.40 \\
MBPP & 68.15 & \textbf{69.92} & 68.91 & 67.61 & \textbf{70.99} & 66.79 \\
MBPP+ & 58.02 & 57.94 & 57.80 & \textbf{59.36} & 55.26 & 55.07 \\
\cellcolor{gray!15}Coding Avg. & \cellcolor{gray!15}61.89 & \cellcolor{gray!15}\textbf{63.70} & \cellcolor{gray!15}62.23 & \cellcolor{gray!15}62.22 & \cellcolor{gray!15}\textbf{63.48} & \cellcolor{gray!15}58.97 \\

\midrule
\multicolumn{7}{c}{\textbf{Math}} \\
\midrule
Math 500 & 73.25 & 74.05 & \textbf{75.50} & 71.50 & 75.60 & \textbf{80.20} \\
GSM8k CoT & \textbf{89.69} & 89.08 & 89.54 & 87.79 & 85.97 & \textbf{87.87} \\
\cellcolor{gray!15}Math Avg. & \cellcolor{gray!15}81.47 & \cellcolor{gray!15}81.57 & \cellcolor{gray!15}\textbf{82.52} & \cellcolor{gray!15}79.65 & \cellcolor{gray!15}80.79 & \cellcolor{gray!15}\textbf{84.04} \\

\midrule
\multicolumn{7}{c}{\textbf{Commonsense Reasoning}} \\
\midrule
Race & \textbf{85.84} & 85.55 & 85.45 & 85.17 & \textbf{86.32} & 85.26 \\
ARC Challenge & 89.16 & \textbf{89.50} & 88.05 & 88.40 & \textbf{89.16} & 89.09 \\
Winogrande & 74.19 & 72.69 & \textbf{74.51} & 72.93 & \textbf{75.37} & 73.95 \\
Hellaswag & 81.23 & 81.39 & \textbf{81.51} & 81.06 & \textbf{82.37} & 82.33 \\
\cellcolor{gray!15}Commonsense Avg. & \cellcolor{gray!15}\textbf{82.61} & \cellcolor{gray!15}82.28 & \cellcolor{gray!15}82.38 & \cellcolor{gray!15}81.89 & \cellcolor{gray!15}\textbf{83.31} & \cellcolor{gray!15}82.66 \\

\bottomrule
\end{tabular}%
}
\caption{Benchmark results across MMLU, Coding, Math, and Commonsense Reasoning for Muon and AdamW optimizers at varying batch sizes ($n\times$GBS) for the Nano-V3 and 8B Hybrid MTP models. 1$\times$GBS is equivalent to 25xM tokens per batch.}
\label{tab:gbs_ablation_results}
\end{table}

% Set width to \linewidth so the image fills the wrapfigure box perfectly

\subsection{Training Stability with SOAP}\label{sec:soap-stability}
We observed significant empirical hurdles when scaling SOAP to large-batch, large parameter count regimes. Here we document the instability with SOAP before moving on to experiments that compare them.

\subsubsection{Diagnosing the "Slingshot" Instability}
\begin{figure}[h]
 \centering
\includegraphics[width=\textwidth, height=0.85\textheight, keepaspectratio]{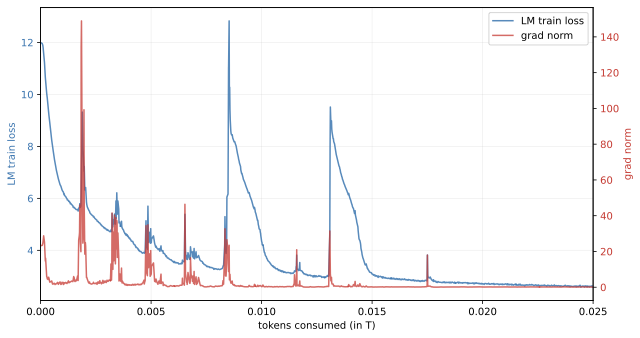}
 \caption{SOAP with a preconditioner that is computed based on stale gradient statistics leads to an oscillatory instability of loss spikes preceded by gradient spikes. The experiment was performed on a dense 621M pure transformer model pretrained with a global batch size of 12.5M tokens.}
 \label{fig_soap_waves}
\end{figure}

The reference SOAP implementation \footnote{\url{https://github.com/nikhilvyas/SOAP}} mitigates the computational expense of QR decomposition by refreshing the eigenbasis infrequently (e.g., every 10 steps) but also simultaneously omits the current step's gradient from the eigenbasis recalculation. While harmless for smaller models and higher eigenbasis recomputation intervals, our large-batch pretraining experiments reveal that this configuration induces severe early-training instability. As illustrated in Fig.~\ref{fig_soap_waves}, this manifests as oscillating weight gradient norms followed closely by spikes in the language modeling loss.

\begin{figure}
 \centering
\includegraphics[width=\linewidth, keepaspectratio]{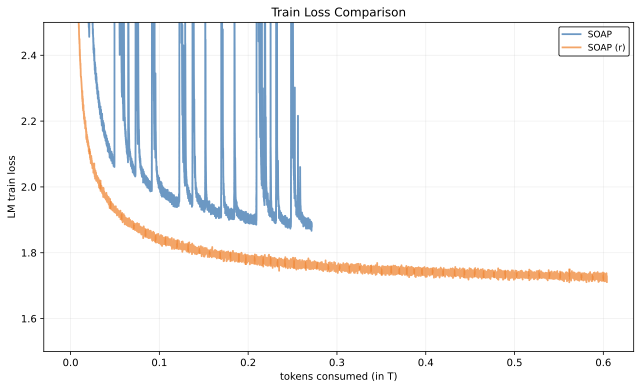}
 \caption{The instability of Fig.~\ref{fig_soap_waves} at a larger model size leads to training divergence in SOAP (blue) and large loss gaps relative to a baseline that uses current gradient statistics to compute preconditioners (in orange). }
 \label{fig_one_off_soap_large}
\end{figure}

We attribute this to a "stale preconditioner" effect. Because the loss landscape changes rapidly during the early phases of large-batch training, applying a preconditioning matrix derived from outdated gradient statistics causes the optimization trajectory to "slingshot". For the smaller 621M model, these oscillations are eventually damped, and the model converges, Fig.~\ref{fig_soap_waves}. However, as we scale up to an 8B dense model, this lag becomes catastrophic, leading to training divergence and significantly poorer quality compared to AdamW, Fig.~\ref{fig_one_off_soap_large}.

  \begin{minipage}{\linewidth}
    \begin{algorithm}[H]
    \caption{KL-SOAP used in experiments, based on~\cite{vyas2024soap, lin2025understanding}}
    \label{soap_algo}
    \begin{algorithmic}[1] 
    \State \textbf{Hyperparameters:} $\beta_{\text{kron}},\, \beta_1,\, \beta_2,\, p,\,  \eta_t,\, \epsilon,\, F,\, \lambda_t$
    
    \State \textbf{Optimizer States:} 
    $
        \mathbf{m}_t ,\, \mathbf{v}_t \in \mathbb{R}^{\text{in}\times \text{out}};\,
        \mathbf{L}_t, \mathbf{Q_L}  \in \mathbb{R}^{\text{in}\times \text{in}};\,
        \mathbf{R}_t, \mathbf{Q_R} \in \mathbb{R}^{\text{out}\times \text{out}}
    $
    \State $\mathbf{G}_t = \nabla_{\mathbf{W}} L(\mathbf{W}_{t-1})$ 
    
    \texttt{Accumulate Kronecker Factors (KL-Shampoo):}
    \State $\mathbf{\lambda}_L = \text{diag}(\mathbf{Q}_L^T \mathbf{L}_{t-1} \mathbf{Q}_L)$ \Comment{Approximate eigenvalues}
    \State $\mathbf{\lambda}_R = \text{diag}(\mathbf{Q}_R^T \mathbf{R}_{t-1} \mathbf{Q}_R)$
    \State $\mathbf{L}_{t} \gets (1-\beta_{\text{kron}})\,\mathbf{L}_{t-1} + \frac{\beta_{\text{kron}}}{\text{out}}\,\mathbf{G}_t\,(\lambda_R+\epsilon)^{-p} \,\mathbf{G}_t^T$
    \State $\mathbf{R}_{t} \gets (1-\beta_{\text{kron}})\,\mathbf{R}_{t-1} + \frac{\beta_{\text{kron}}}{\text{in}}\,\mathbf{G}_t^T\,(\lambda_L+\epsilon)^{-p} \,\mathbf{G}_t$
    
    \texttt{Eigenbasis Re-computation}
    \If{$t \bmod F = 0$}
        \State $\hat{\mathbf{m}}_t = \mathbf{Q}_L\,\mathbf{m}_t\,\mathbf{Q}_R^T $
        \State $\mathbf{Q}_{L}\,_ = \texttt{QR}(\mathbf{L}_{t}\,\mathbf{Q}_{L}).Q$
        \State $\mathbf{Q}_{R}\,_ = \texttt{QR}(\mathbf{R}_{t}\,\mathbf{Q}_{R}).Q$
        \State $\mathbf{m}_t = \mathbf{Q}_L^T\,\hat{\mathbf{m}}_t\,\mathbf{Q}_R$
    \EndIf
    
    \texttt{Rotate gradient into eigenbasis}
    \State $\mathbf{G}_t' \gets \hat{\mathbf{Q}}_L^T\,\mathbf{G}_t\,\hat{\mathbf{Q}}_R$
    
    \texttt{Calculate Adam update}
    \State $\mathbf{m}_t \gets (1 - \beta_1) \, \mathbf{m}_{t-1} + \beta_1\,\mathbf{G}'$
    \State $\mathbf{v}_t \gets (1 - \beta_2)\,\mathbf{v}_{t-1} + \beta_2\,\mathbf{G}_t'\odot \mathbf{G}_t'$
    \State $\mathbf{N} \gets \dfrac{\mathbf{m}_t}{\sqrt{\mathbf{v}_t}+\epsilon}$ 
    
    \texttt{Rotate update out of eigenbasis}
    \State $\mathbf{\Delta W}=\mathbf{Q}_L\,\mathbf{N}\,\mathbf{Q}_R^T$
    
    \texttt{Weight decay and update}
    \State $\mathbf{W}_{t} \gets (1-\eta_t\lambda_t)\mathbf{W}_{t-1} - \eta_t\mathbf{\Delta W}$
    
    \end{algorithmic}
    \end{algorithm}
  \end{minipage}

\begin{table}[h]
\centering
\small
\begin{tabular}{lll}
\toprule
\textbf{Hyperparameter} & \textbf{Description} & \textbf{Value} \\
\midrule
$\beta_{\text{kron}}$ & EMA coefficient for accumulating Kronecker factors & 0.95 \\
$\beta_1$ & First-moment EMA coefficient in the rotated basis & 0.9 \\
$\beta_2$ & Second-moment EMA coefficient in the rotated basis & 0.95 \\
$p$ & Exponent used in the KL-Shampoo covariance update & -1 \\
$F$ & Eigenbasis recomputation frequency & 1 \\
$\lambda_t$ & Decoupled weight decay coefficient & 0.1 \\
$\epsilon$ & Numerical stability constant for KL-covariance and Adam updates & $10^{-8}$ \\
\bottomrule
\end{tabular}
\caption{KL-SOAP hyperparameters used in Algorithm~\ref{soap_algo}.}
\label{tab:soap_hparams}
\end{table}

\subsubsection{Mitigating Staleness: Real-Time Eigenbasis Updates}

We find that simply increasing the eigenbasis update frequency to every step is insufficient if the current iteration's gradient is excluded. It is the combination of per-step QR updates \textit{and} the inclusion of the current step’s gradient in the eigenbasis calculation that fully resolves the issue. When modified to use real-time gradient statistics, SOAP eliminates the loss spikes and achieves an LM loss on par with Muon, Fig.~\ref{fig:muon_vs_soap_comparative}. We note that this implementation was present with SOAP and is not present in \cite{anil2020scalable, lin2025understanding} and other works.

\subsubsection{Gradient Covariance Estimation via Kullback–Leibler Minimization}
\begin{figure}[h] 
\centering\includegraphics[scale=0.5]{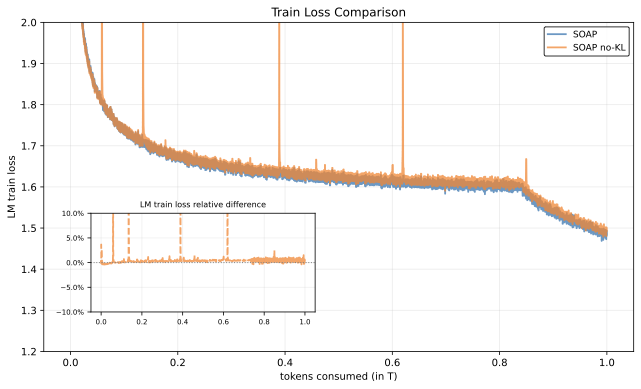} 
\caption{The inclusion of KL-divergence-based gradient covariance estimation \cite{lin2025understanding} in SOAP stabilizes training and results in lower training loss and fewer loss spikes. Experiments were performed on a Qwen-3-30B-A3B architecture.}
\label{fig_unstable_large}
\end{figure}

Even with real-time updates, the accumulation of Kronecker factor matrices ($GG^T$ and $G^TG$) can suffer from instabilities at scale. As described in \cite{lin2025understanding}, we find that replacing the standard accumulation with the KL divergence-regularized covariance estimation algorithm  provides a critical secondary layer of stability. As shown in Fig.~\ref{fig_unstable_large}, integrating KL-Shampoo's update rule further suppresses loss spikes and yields a lower final training loss on the Qwen-3-30B-A3B architecture.

This empirical improvement comes from better estimation of the gradient covariance, and the stability benefit comes from the spectral properties of the update. As detailed in Appendix \ref{appendex:kl_condition}, KL-Shampoo reduces the condition number of the Kronecker factor matrices to the square root of the standard formulation:
\begin{equation}\kappa(\mathbf{S}_{\text{KL-Shampoo}}) = \sqrt{\kappa(\mathbf{S}_{\text{Shampoo}})}.\end{equation}
We hypothesize that this property improves the numerical stability of the eigenbasis recomputation, whether using QR or eigendecomposition-based eigensolvers, safeguarding the training dynamics against floating-point noise and preventing subsequent gradient explosions or loss spikes.

\subsection{Comparing Muon and SOAP}\label{sec:compare-muon-soap}

To evaluate the relative performance of Muon and SOAP, we use a controlled comparison. In both configurations, we employ the simplest "blocking" method for tensor parallelism support. Further, to ensure an apples-to-apples structural comparison, we disable Query-Key-Value (QKV) splitting in Muon. By treating the fused QKV projection as a single matrix, matching the default operational mode of our implementation of SOAP, we isolate the algorithmic differences in preconditioning and orthogonalization from structural choices.

\begin{figure}[htbp]
    \centering
    % First figure (Top)
    \includegraphics[width=0.65\linewidth]{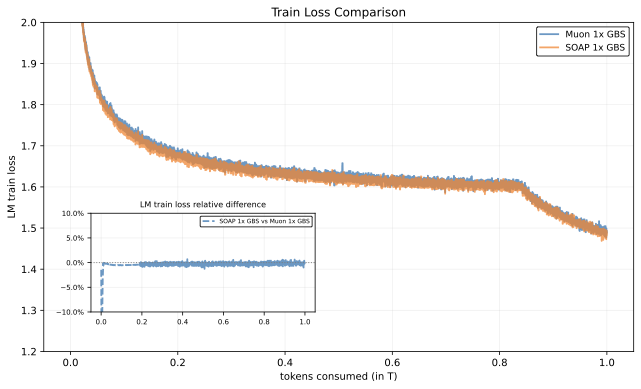}
    
    \vspace{2ex} % Adds vertical space between the images
    
    % Second figure (Middle)
    \includegraphics[width=0.65\linewidth]{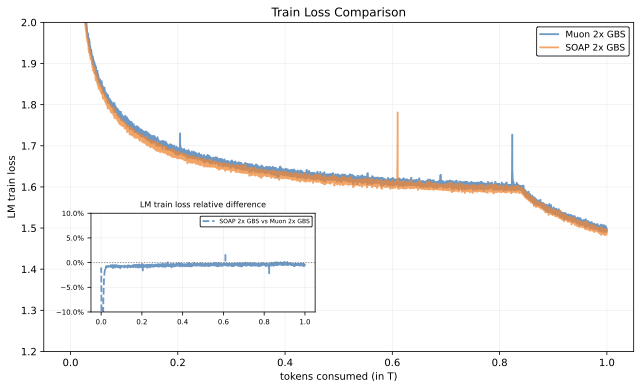}
    
    \vspace{2ex} % Adds vertical space between the images
    
    % Third figure (Bottom)
    \includegraphics[width=0.65\linewidth]{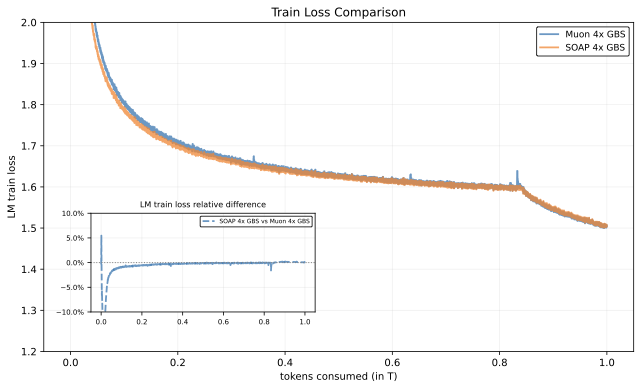}
    
    \caption{Large-batch comparative study of Muon and SOAP for the Qwen3-30B-A3B architecture (Top: 1x, Middle: 2x, Bottom: 4x GBS). Inset shows relative loss differences. Note that these experiments did not use any batch-size ramp-up.}
    \label{fig:muon_vs_soap_comparative}
\end{figure}

\begin{figure}
    \centering
\includegraphics[width=0.85\textwidth]{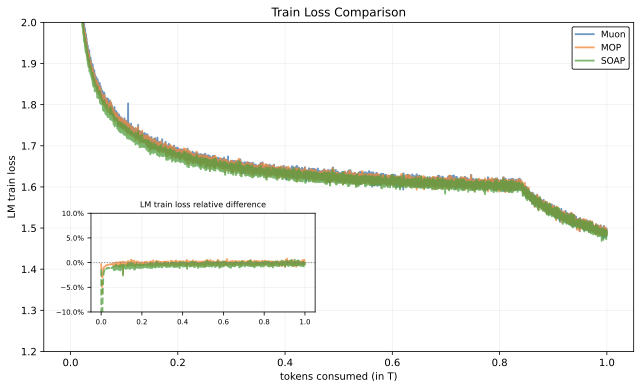}
    \caption{Qwen-3-30B-A3B architecture pretrained with Muon, SOAP, and MOP with update-RMS matching. Both SOAP and MOP have slightly lower losses than Muon. Inset shows relative loss differences to the Muon baseline. The global batch size was fixed to 24M tokens for this experiment.}
    \label{fig:muon_mop_soap_compare}
\end{figure}

We do a comparative analysis for MoEs using the Qwen-3-30B-A3B architecture. Fig. ~\ref{fig:muon_vs_soap_comparative} shows the results of a large batch study comparing Muon and SOAP. We also include a version of Muon we term Momentum Orthogonalized by Polar (MOP), which is Muon that replaces Newton-Schulz orthogonalization with exact polar decomposition via SVD \footnote{This was first developed in \url{https://github.com/KellerJordan/cifar10-airbench/tree/master}}. Fig.~\ref{fig:muon_mop_soap_compare} presents the training trajectories for these optimizers under the same controlled tensor parallelism and non-split QKV conditions. Throughout the majority of the pretraining run, KL-SOAP maintains a consistent, albeit slight, edge over Muon in the CE loss.

\paragraph{Tuning the "Epsilon"}\label{sec:tuning-epsilon}

Both SOAP and Muon expose an $\epsilon$ hyperparameter, and recent work has shown that this seemingly small numerical constant can affect optimizer stability, transfer across scale, and the predictability of training dynamics \cite{everett2024scaling, wortsman2023small, yang2023tensor, epsilon_nuisance}. In SOAP, $\epsilon$ plays a role analogous to the $\epsilon$ term in AdamW: it floors the denominator of the second-moment estimate after the momentum has been rotated into the preconditioned eigenbasis and the eigenvalues in the KL-Shampoo accumulation step. In addition, because SOAP uses Shampoo-style kronecker factors where the preconditioning is effectively applying inverse square powers to the kronecker factor eigenvalues \cite{vyas2024soap}, the $\epsilon$ used for the Adam-like computation in the preconditioner basis can be interpreted as setting a soft floor on smallest effective eigenvalue of the Shampoo preconditioner.
In Muon, $\epsilon$ has a different role and is used directly in the normalization step before Newton-Schulz orthogonalization, where the momentum matrix is scaled by $\max(\|\mathbf{M}_t\|_F,\epsilon)$. Thus, Muon's $\epsilon$ lower-bounds the Frobenius norm used for matrix normalization rather than flooring an element-wise second-moment estimate. We therefore treat $\epsilon$ as an optimizer-specific hyperparameter: for SOAP it controls the numerical floor in preconditioned adaptive moments and factor inversions, while for Muon it controls the smallest normalization scale entering the Newton-Schulz iteration.

\paragraph{Limitations of our work}In this report, we do not tune $\epsilon$ systematically. We use $\epsilon=10^{-8}$ for SOAP and AdamW, and $\epsilon=10^{-7}$ for Muon. Understanding how these values should scale with model size, optimizer parameterization, precision format, and training horizon is an important direction for subsequent work.

\section{Implementation Details: Layer-Wise Distributed Optimizer}\label{sec:layerwise-optimizer}
While higher-order optimizers like SOAP and Muon demonstrate better convergence and stability at larger batch sizes, realizing their empirical benefits in production environments requires the underlying training system to supply the optimizers with unfragmented 2D weight matrices without incurring too much exposed communication overhead. This conflicts with the tensor sharding inherent to standard distributed training approaches (e.g., ZeRO\cite{rajbhandari2020zero} or FSDP\cite{zhao2023pytorch}). To overcome this system bottleneck, we developed a strategy native to the Megatron-LM framework \cite{shoeybi2019megatron}. 

This implementation outlines a distributed optimizer strategy designed to support matrix-based optimizers by keeping parameter matrices fully intact, while still balancing memory and hiding communication bottlenecks. Here we describe the main features of the implementation, which is available in Megatron-LM \footnote{\url{https://github.com/NVIDIA/Megatron-LM/blob/main/megatron/core/optimizer/layer_wise_optimizer.py}}. 
% it is an evolving implementation
% there are x problems making distribution of muon hard, we make the following choices to address them partially now, our future implementations will improve these choices.
\begin{enumerate}
\item Load Balancing: Instead of flattening and slicing individual tensors, entire parameter matrices are sorted by size and distributed across GPUs in a round robin sequence. This ensures every GPU has a roughly equal memory load without fragmenting the matrices. 
%\item Gradient Communication: Gradients are computed and fully synchronized across all GPUs so every GPU temporarily holds the full model's gradients.
\item Parameter Update: Each GPU acts independently to update only its assigned parameters. It applies the optimizer math (e.g., using Muon or SOAP for 2D matrices and AdamW for standard vectors) to its specific layers, and then flattens the results into a buffer to be gathered in the next step's forward pass.
% it eliminates redudant computation and is immediately paralleizable
\item Overlapped Parameter All-Gather: Parameter synchronization is chunked into sequential communication buckets that logically align with the model's execution order (leveraging Megatron's DDP bucket infrastructure). By grouping layers into these buckets, the system bounds message sizes and enables pipelining. Instead of a single, monolithic blocking synchronization, the system issues an asynchronous all-gather collective for each bucket. As the forward pass computes activations using the current bucket's weights, the GPUs simultaneously execute the all-gather to fetch the updated parameter matrices for the upcoming bucket from their respective owners, effectively hiding the network latency behind the computation. Keeping parameter matrices intact results in slightly uneven total parameter counts across GPUs, so standard uniform all-gather collectives cannot be used. Instead, the asynchronous synchronization employs a variable-sized allgather-V \footnote{\url{https://www.mpich.org/static/docs/v3.2/www3/MPI_Allgatherv.html}} collective. This allows each GPU to synchronize its uniquely sized, contiguous buffer of updated layers to the rest of the ranks natively, avoiding the need for memory-wasting padding or tensor fragmentation.
\end{enumerate}

\section{Conclusions and Future Work}

In this report, we conducted large-scale pretraining studies of the Muon and SOAP optimizers, identifying key best practices for maintaining stability at large batch sizes. We have open-sourced our implementations in Megatron-LM and \texttt{Emerging-Optimizers}. Our findings demonstrate that both SOAP and Muon consistently outperform AdamW, scaling effectively to significantly larger batch sizes during pretraining. Furthermore, the KL-SOAP variant emerged as the most effective approach overall; therefore, in scenarios where memory footprint is not a limiting factor, we recommend KL-SOAP over Muon.

We summarize several algorithmic, system-level, and empirical directions for advancing and scaling these research directions below.

\paragraph{SOAP}
To fully realize the benefits of SOAP at frontier model scales, it is essential to support additional structural and distributed features. Specifically, extending our layer-wise distributed optimizer to natively support Tensor Parallelism (TP) for SOAP, handling fused tensors, for example by implementing QKV splitting for attention, and further optimizing the throughput of optimizer matrix multiplications such as the KL-Shampoo eigenbasis updates are areas of potential improvement. 
%Additionally, because SOAP maintains significantly larger optimizer states than AdamW or Muon, developing robust memory management and synchronization strategies such as asynchronous host-device (CPU) state offloading remains a critical priority for future scaling.

\paragraph{Optimizer-Driven Aligned Memory Buffer Layouts}
Currently, standard Distributed Data Parallel (DDP) implementations force parameters into rigid, 1D contiguous memory buffers. While our layer-wise algorithm successfully navigates this via bucketed, variable-sized collectives, the underlying buffer allocation remains largely optimizer-agnostic. This creates inherent friction for higher-order methods like SOAP and Muon, which operate strictly on 2D matrices and structured factorization states.
A direction for future work is to refactor the DDP buffer allocation engine to be natively optimizer-driven. Rather than having the DDP module flatten parameters without considering the structure and boundaries of network weights, the system should allow the optimizer to pre-compute and dictate the layout of the parameter and gradient buffers. By granting the optimizer control, the physical communication buckets will map to the structural shape of the optimizer computations, reducing communication overhead. This optimization is being built in the Megatron-LM framework \footnote{\url{https://github.com/NVIDIA/Megatron-LM/pull/4509}} and will replace gradient all-reduces with reduce-scatters and further optimize memory management and sharding.

\paragraph{Optimizer-Aware Architectural Co-Design}
We observed that applying orthogonalization to Mamba2's Conv1D filters degraded accuracy and sometimes caused instability in the form of NaNs, highlighting that matrix-based optimizers can be geometrically mismatched with certain architectures. Further investigation is needed to map out how architectural components such as Multi-head Latent Attention (MLA) \cite{liu2024deepseek} or low-rank adapters (LoRA) interact with full-rank preconditioning assumptions, potentially leading to hybrid training recipes that selectively route parameters to AdamW or Muon or a variant based on their tensor shapes and operator functions \cite{bernstein2024modular}. Along these lines, recent work has developed a matrix-based optimizer for native low-rank factored training \cite{janson2026stabilizing}.
\paragraph{Large Batch Training and Scaling Laws for Batch Size}
Recent empirical studies have established scaling laws to predict optimal batch sizes as a function of compute budget and training hyperparameters~\cite{bi2024deepseek, li2025minimax, li2025predictable}, with theoretical foundations linking these scaling laws to the interplay between gradient noise reduction and optimization step efficiency~\cite{islamov2026role}. While we do not explicitly derive or measure scaling laws in this work, we instead approach batch size primarily as a systems-level constraint. Specifically, we push the global batch size as high as possible, while maintaining quality, with the goal of maximizing GPU utilization by minimizing data-parallel communication overhead, which is necessary for efficient pretraining at scale. An optimal approach would combine these two points of view, for example, by using scaling laws to better tune the batch size ramp-up and learning rate schedule and heuristics for changing the maximum batch size as token horizons vary.

\paragraph{More Accurate Orthogonalization at Larger Scales}
Scaling matrix-based optimizers that perform orthogonalization to larger models and longer token horizons can introduce new numerical challenges that warrant further investigation. The typical size of matrices increases as we scale models up, and so does the ``heavy-tail" of small singular values of these matrices' gradients. Since the optimizer step always occurs in FP32 single precision, this ``heavy-tail" of small singular values and their associated vectors is sensitive to numerical issues and can become increasingly susceptible to floating-point noise, often being below FP32 machine precision ($2^{-23}\approx 1.19\times 10^{-7}$). For Muon's Newton-Schulz iterations which occur in BF16, it is coarser ($2^{-7} = 0.0078125$) and small singular values often represent numerical noise rather than informative signal, creating an effective "noise floor" in FP32. Truncated or regularized orthogonalization approaches and corresponding hardware-friendly algorithms that implement them are open research directions, with notable recent work measuring spectral scaling laws for these quantities~\cite{magakyan2026spectral}.

\newpage
\bibliographystyle{unsrt}
\bibliography{bib}

@article{vyas2024soap,
  title={Soap: Improving and stabilizing shampoo using adam},
  author={Vyas, Nikhil and Morwani, Depen and Zhao, Rosie and Kwun, Mujin and Shapira, Itai and Brandfonbrener, David and Janson, Lucas and Kakade, Sham},
  journal={arXiv preprint arXiv:2409.11321},
  year={2024}
}

@article{eschenhagen2025purifying,
  title={Purifying Shampoo: Investigating Shampoo's Heuristics by Decomposing its Preconditioner},
  author={Eschenhagen, Runa and Defazio, Aaron and Lee, Tsung-Hsien and Turner, Richard E and Shi, Hao-Jun Michael},
  journal={arXiv preprint arXiv:2506.03595},
  year={2025}
}

@article{anil2020scalable,
  title={Scalable second order optimization for deep learning},
  author={Anil, Rohan and Gupta, Vineet and Koren, Tomer and Regan, Kevin and Singer, Yoram},
  journal={arXiv preprint arXiv:2002.09018},
  year={2020}
}

@inproceedings{gupta2018shampoo,
  title={Shampoo: Preconditioned stochastic tensor optimization},
  author={Gupta, Vineet and Koren, Tomer and Singer, Yoram},
  booktitle={International Conference on Machine Learning},
  pages={1842--1850},
  year={2018},
  organization={PMLR}
}

@article{morwani2024new,
  title={A New Perspective on Shampoo's Preconditioner},
  author={Morwani, Depen and Shapira, Itai and Vyas, Nikhil and Malach, Eran and Kakade, Sham and Janson, Lucas},
  journal={arXiv preprint arXiv:2406.17748},
  year={2024}
}

@article{jordanmuon,
  title={Muon: An optimizer for hidden layers in neural networks, 2024b},
  author={Jordan, K and Jin, Y and Boza, V and Jiacheng, Y and Cecista, F and Newhouse, L and Bernstein, J},
  journal={URL https://kellerjordan. github. io/posts/muon}
}

@article{bernstein2024modular,
  title={Modular duality in deep learning},
  author={Bernstein, Jeremy and Newhouse, Laker},
  journal={arXiv preprint arXiv:2410.21265},
  year={2024}
}

@article{bernstein2024old,
  title={Old optimizer, new norm: An anthology},
  author={Bernstein, Jeremy and Newhouse, Laker},
  journal={arXiv preprint arXiv:2409.20325},
  year={2024}
}

@article{large2024scalable,
  title={Scalable optimization in the modular norm},
  author={Large, Tim and Liu, Yang and Huh, Minyoung and Bahng, Hyojin and Isola, Phillip and Bernstein, Jeremy},
  journal={arXiv preprint arXiv:2405.14813},
  year={2024}
}

@article{shah2025practical,
  title={Practical efficiency of muon for pretraining},
  author={Shah, Ishaan and Polloreno, Anthony M and Stratos, Karl and Monk, Philip and Chaluvaraju, Adarsh and Hojel, Andrew and Ma, Andrew and Thomas, Anil and Tanwer, Ashish and Shah, Darsh J and others},
  journal={arXiv preprint arXiv:2505.02222},
  year={2025}
}

@article{liu2025muon,
  title={Muon is scalable for llm training},
  author={Liu, Jingyuan and Su, Jianlin and Yao, Xingcheng and Jiang, Zhejun and Lai, Guokun and Du, Yulun and Qin, Yidao and Xu, Weixin and Lu, Enzhe and Yan, Junjie and others},
  journal={arXiv preprint arXiv:2502.16982},
  year={2025}
}

@article{blakeman2025nemotron,
  title={Nemotron-h: A family of accurate and efficient hybrid mamba-transformer models},
  author={Blakeman, Aaron and Basant, Aarti and Khattar, Abhinav and Renduchintala, Adithya and Bercovich, Akhiad and Ficek, Aleksander and Bjorlin, Alexis and Taghibakhshi, Ali and Deshmukh, Amala Sanjay and Mahabaleshwarkar, Ameya Sunil and others},
  journal={arXiv preprint arXiv:2504.03624},
  year={2025}
}

@article{lin2025understanding,
  title={Understanding and improving the shampoo optimizer via kullback-leibler minimization},
  author={Lin, Wu and Lowe, Scott C and Dangel, Felix and Eschenhagen, Runa and Xu, Zikun and Grosse, Roger B},
  journal={arXiv e-prints},
  pages={arXiv--2509},
  year={2025}
}

@article{kexuefm-11307,
  title={{A}dam{W} {W}eight {RMS}},
  author={Jianlin Su},
  year={2025},
  month={Oct},
  journal={\url{https://kexue.fm/archives/11307}},
}

@article{kosson2023rotational,
  title={Rotational equilibrium: How weight decay balances learning across neural networks},
  author={Kosson, Atli and Messmer, Bettina and Jaggi, Martin},
  journal={arXiv preprint arXiv:2305.17212},
  year={2023}
}

@article{li2024stochastic,
  title={Stochastic Hessian Fittings with Lie Groups},
  author={Li, Xi-Lin},
  journal={arXiv preprint arXiv:2402.11858},
  year={2024}
}

@inproceedings{balles2018dissecting,
  title={Dissecting adam: The sign, magnitude and variance of stochastic gradients},
  author={Balles, Lukas and Hennig, Philipp},
  booktitle={International Conference on Machine Learning},
  pages={404--413},
  year={2018},
  organization={PMLR}
}

@article{li2025normuon,
  title={NorMuon: Making Muon more efficient and scalable},
  author={Li, Zichong and Liu, Liming and Liang, Chen and Chen, Weizhu and Zhao, Tuo},
  journal={arXiv preprint arXiv:2510.05491},
  year={2025}
}

@article{si2025adamuon,
  title={Adamuon: Adaptive muon optimizer},
  author={Si, Chongjie and Zhang, Debing and Shen, Wei},
  journal={arXiv preprint arXiv:2507.11005},
  year={2025}
}

@article{xie2026controlled,
  title={Controlled llm training on spectral sphere},
  author={Xie, Tian and Luo, Haoming and Tang, Haoyu and Hu, Yiwen and Liu, Jason Klein and Ren, Qingnan and Wang, Yang and Zhao, Wayne Xin and Yan, Rui and Su, Bing and others},
  journal={arXiv preprint arXiv:2601.08393},
  year={2026}
}

@article{loshchilov2017decoupled,
  title={Decoupled weight decay regularization},
  author={Loshchilov, Ilya and Hutter, Frank},
  journal={arXiv preprint arXiv:1711.05101},
  year={2017}
}

@article{pethick2025training,
  title={Training deep learning models with norm-constrained lmos},
  author={Pethick, Thomas and Xie, Wanyun and Antonakopoulos, Kimon and Zhu, Zhenyu and Silveti-Falls, Antonio and Cevher, Volkan},
  journal={arXiv preprint arXiv:2502.07529},
  year={2025}
}

@article{liu2024deepseek,
  title={Deepseek-v3 technical report},
  author={Liu, Aixin and Feng, Bei and Xue, Bing and Wang, Bingxuan and Wu, Bochao and Lu, Chengda and Zhao, Chenggang and Deng, Chengqi and Zhang, Chenyu and Ruan, Chong and others},
  journal={arXiv preprint arXiv:2412.19437},
  year={2024}
}

@article{mccandlish2018empirical,
  title={An empirical model of large-batch training},
  author={McCandlish, Sam and Kaplan, Jared and Amodei, Dario and Team, OpenAI Dota},
  journal={arXiv preprint arXiv:1812.06162},
  year={2018}
}

@article{sun2024hunyuan,
  title={Hunyuan-large: An open-source moe model with 52 billion activated parameters by tencent},
  author={Sun, Xingwu and Chen, Yanfeng and Huang, Yiqing and Xie, Ruobing and Zhu, Jiaqi and Zhang, Kai and Li, Shuaipeng and Yang, Zhen and Han, Jonny and Shu, Xiaobo and others},
  journal={arXiv preprint arXiv:2411.02265},
  year={2024}
}

@article{george2018fast,
  title={Fast approximate natural gradient descent in a kronecker factored eigenbasis},
  author={George, Thomas and Laurent, C{\'e}sar and Bouthillier, Xavier and Ballas, Nicolas and Vincent, Pascal},
  journal={Advances in neural information processing systems},
  volume={31},
  year={2018}
}

@inproceedings{martens2015optimizing,
  title={Optimizing neural networks with kronecker-factored approximate curvature},
  author={Martens, James and Grosse, Roger},
  booktitle={International conference on machine learning},
  pages={2408--2417},
  year={2015},
  organization={PMLR}
}

@inproceedings{bollapragada2018progressive,
  title={A progressive batching L-BFGS method for machine learning},
  author={Bollapragada, Raghu and Nocedal, Jorge and Mudigere, Dheevatsa and Shi, Hao-Jun and Tang, Ping Tak Peter},
  booktitle={International Conference on Machine Learning},
  pages={620--629},
  year={2018},
  organization={PMLR}
}

@article{modoranu2026dash,
  title={DASH: Faster Shampoo via Batched Block Preconditioning and Efficient Inverse-Root Solvers},
  author={Modoranu, Ionut-Vlad and Zmushko, Philip and Schultheis, Erik and Safaryan, Mher and Alistarh, Dan},
  journal={arXiv preprint arXiv:2602.02016},
  year={2026}
}

@online{wen2025hyperball_part1,
    title        = {Fantastic Pretraining Optimizers and Where to Find Them II: From Weight Decay to Hyperball Optimization},
    author       = {Wen, Kaiyue and Dang, Xingyu and Lyu, Kaifeng and Ma, Tengyu and Liang, Percy},
    year         = {2025},
    month        = {12},
    day          = {15},
    url          = {https://whenwen.github.io/wd_blog/public/hyperball-part-1.html},
    urldate      = {2025-12-15},
}

@inproceedings{carlson2015stochastica,
  title={Stochastic spectral descent for restricted Boltzmann machines},
  author={Carlson, David and Cevher, Volkan and Carin, Lawrence},
  booktitle={Artificial intelligence and statistics},
  pages={111--119},
  year={2015},
  organization={PMLR}
}

@article{carlson2015stochasticb,
  title={Stochastic spectral descent for discrete graphical models},
  author={Carlson, David and Hsieh, Ya-Ping and Collins, Edo and Carin, Lawrence and Cevher, Volkan},
  journal={IEEE Journal of Selected Topics in Signal Processing},
  volume={10},
  number={2},
  pages={296--311},
  year={2015},
  publisher={IEEE}
}

@article{carlson2015preconditioned,
  title={Preconditioned spectral descent for deep learning},
  author={Carlson, David E and Collins, Edo and Hsieh, Ya-Ping and Carin, Lawrence and Cevher, Volkan},
  journal={Advances in neural information processing systems},
  volume={28},
  year={2015}
}

@article{flynn2017duality,
  title={The duality structure gradient descent algorithm: analysis and applications to neural networks},
  author={Flynn, Thomas},
  journal={arXiv preprint arXiv:1708.00523},
  year={2017}
}

@inproceedings{rajbhandari2020zero,
  title={Zero: Memory optimizations toward training trillion parameter models},
  author={Rajbhandari, Samyam and Rasley, Jeff and Ruwase, Olatunji and He, Yuxiong},
  booktitle={SC20: international conference for high performance computing, networking, storage and analysis},
  pages={1--16},
  year={2020},
  organization={IEEE}
}

@article{zhao2023pytorch,
  title={Pytorch fsdp: experiences on scaling fully sharded data parallel},
  author={Zhao, Yanli and Gu, Andrew and Varma, Rohan and Luo, Liang and Huang, Chien-Chin and Xu, Min and Wright, Less and Shojanazeri, Hamid and Ott, Myle and Shleifer, Sam and others},
  journal={arXiv preprint arXiv:2304.11277},
  year={2023}
}

@article{amsel2025polar,
  title={The polar express: Optimal matrix sign methods and their application to the muon algorithm},
  author={Amsel, Noah and Persson, David and Musco, Christopher and Gower, Robert M},
  journal={arXiv preprint arXiv:2505.16932},
  year={2025}
}

@article{brown2020language,
  title={Language models are few-shot learners},
  author={Brown, Tom and Mann, Benjamin and Ryder, Nick and Subbiah, Melanie and Kaplan, Jared D and Dhariwal, Prafulla and Neelakantan, Arvind and Shyam, Pranav and Sastry, Girish and Askell, Amanda and others},
  journal={Advances in neural information processing systems},
  volume={33},
  pages={1877--1901},
  year={2020}
}

@article{yang2025qwen3,
  title={Qwen3 technical report},
  author={Yang, An and Li, Anfeng and Yang, Baosong and Zhang, Beichen and Hui, Binyuan and Zheng, Bo and Yu, Bowen and Gao, Chang and Huang, Chengen and Lv, Chenxu and others},
  journal={arXiv preprint arXiv:2505.09388},
  year={2025}
}

@article{li2025minimax,
  title={Minimax-01: Scaling foundation models with lightning attention},
  author={Li, Aonian and Gong, Bangwei and Yang, Bo and Shan, Boji and Liu, Chang and Zhu, Cheng and Zhang, Chunhao and Guo, Congchao and Chen, Da and Li, Dong and others},
  journal={arXiv preprint arXiv:2501.08313},
  year={2025}
}

@inproceedings{You2020Large,
    title={Large Batch Optimization for Deep Learning: Training BERT in 76 minutes},
    author={Yang You and Jing Li and Sashank Reddi and Jonathan Hseu and Sanjiv Kumar and Srinadh Bhojanapalli and Xiaodan Song and James Demmel and Kurt Keutzer and Cho-Jui Hsieh},
    booktitle={International Conference on Learning Representations},
    year={2020},
    url={https://openreview.net/forum?id=Syx4wnEtvH}
}

@inproceedings{hoffer2017train,
    author = {Hoffer, Elad and Hubara, Itay and Soudry, Daniel},
    title = {Train longer, generalize better: closing the generalization gap in large batch training of neural networks},
    year = {2017},
    address = {Red Hook, NY, USA},
    booktitle = {Proceedings of the 31st International Conference on Neural Information Processing Systems},
    series = {NIPS'17}
}

@article{you2017large,
  title={Large batch training of convolutional networks},
  author={You, Yang and Gitman, Igor and Ginsburg, Boris},
  journal={arXiv preprint arXiv:1708.03888},
  year={2017}
}

@article{krizhevsky2014one,
  title={One weird trick for parallelizing convolutional neural networks},
  author={Krizhevsky, Alex},
  journal={arXiv preprint arXiv:1404.5997},
  year={2014}
}

@inproceedings{gloeckle2024better,
    author = {Gloeckle, Fabian and Idrissi, Badr Youbi and Rozi\`{e}re, Baptiste and Lopez-Paz, David and Synnaeve, Gabriel},
    title = {Better \& faster large language models via multi-token prediction},
    year = {2024},
    booktitle = {Proceedings of the 41st International Conference on Machine Learning},
    series = {ICML'24}
}

@article{wang2026vescale,
  title={veScale-FSDP: Flexible and High-Performance FSDP at Scale},
  author={Wang, Zezhou and Li, Youjie and Lin, Zhiqi and Yang, Jiacheng and Xie, Cong and Feng, Guanyu and Zhong, Zheng and Huang, Ziyue and Zhu, Hongyu and Zhang, Zhi and others},
  journal={arXiv preprint arXiv:2602.22437},
  year={2026}
}

@article{wang2026canzona,
  title={Canzona: A Unified, Asynchronous, and Load-Balanced Framework for Distributed Matrix-based Optimizers},
  author={Wang, Liangyu and Zhang, Siqi and Wang, Junjie and Dong, Yiming and Zheng, Bo and Qiu, Zihan and Tang, Shengkun and Wang, Di and Men, Rui and Liu, Dayiheng},
  journal={arXiv preprint arXiv:2602.06079},
  year={2026}
}

@article{goyal2017accurate,
  title={Accurate, large minibatch sgd: Training imagenet in 1 hour},
  author={Goyal, Priya and Doll{\'a}r, Piotr and Girshick, Ross and Noordhuis, Pieter and Wesolowski, Lukasz and Kyrola, Aapo and Tulloch, Andrew and Jia, Yangqing and He, Kaiming},
  journal={arXiv preprint arXiv:1706.02677},
  year={2017}
}

@article{dao2024transformers,
  title={Transformers are ssms: Generalized models and efficient algorithms through structured state space duality},
  author={Dao, Tri and Gu, Albert},
  journal={arXiv preprint arXiv:2405.21060},
  year={2024}
}

@article{hagele2024scaling,
  title={Scaling laws and compute-optimal training beyond fixed training durations},
  author={H{\"a}gele, Alexander and Bakouch, Elie and Kosson, Atli and Allal, Loubna B and Von Werra, Leandro and Jaggi, Martin},
  journal={Advances in Neural Information Processing Systems},
  volume={37},
  pages={76232--76264},
  year={2024}
}

@article{mishra2025recipes,
  title={Recipes for Pre-training LLMs with MXFP8},
  author={Mishra, Asit and Stosic, Dusan and Layton, Simon and Micikevicius, Paulius},
  journal={arXiv preprint arXiv:2506.08027},
  year={2025}
}

@article{chen2023symbolic,
  title={Symbolic discovery of optimization algorithms},
  author={Chen, Xiangning and Liang, Chen and Huang, Da and Real, Esteban and Wang, Kaiyuan and Pham, Hieu and Dong, Xuanyi and Luong, Thang and Hsieh, Cho-Jui and Lu, Yifeng and others},
  journal={Advances in neural information processing systems},
  volume={36},
  pages={49205--49233},
  year={2023}
}

@article{janson2026stabilizing,
  title={Stabilizing Native Low-Rank LLM Pretraining},
  author={Janson, Paul and Oyallon, Edouard and Belilovsky, Eugene},
  journal={arXiv preprint arXiv:2602.12429},
  year={2026}
}

@article{malladi2022sdes,
  title={On the SDEs and scaling rules for adaptive gradient algorithms},
  author={Malladi, Sadhika and Lyu, Kaifeng and Panigrahi, Abhishek and Arora, Sanjeev},
  journal={Advances in Neural Information Processing Systems},
  volume={35},
  pages={7697--7711},
  year={2022}
}

@article{bi2024deepseek,
  title={Deepseek llm: Scaling open-source language models with longtermism},
  author={Bi, Xiao and Chen, Deli and Chen, Guanting and Chen, Shanhuang and Dai, Damai and Deng, Chengqi and Ding, Honghui and Dong, Kai and Du, Qiushi and Fu, Zhe and others},
  journal={arXiv preprint arXiv:2401.02954},
  year={2024}
}

@article{li2025predictable,
  title={Predictable Scale: Part I, Step Law--Optimal Hyperparameter Scaling Law in Large Language Model Pretraining},
  author={Li, Houyi and Zheng, Wenzhen and Wang, Qiufeng and Zhang, Hanshan and Wang, Zili and Xuyang, Shijie and Fan, Yuantao and Ding, Zhenyu and Wang, Haoying and Ding, Ning and others},
  journal={arXiv preprint arXiv:2503.04715},
  year={2025}
}

@article{islamov2026role,
  title={On the Role of Batch Size in Stochastic Conditional Gradient Methods},
  author={Islamov, Rustem and Machacek, Roman and Lucchi, Aurelien and Silveti-Falls, Antonio and Gorbunov, Eduard and Cevher, Volkan},
  journal={arXiv preprint arXiv:2603.21191},
  year={2026}
}

@article{shoeybi2019megatron,
  title={Megatron-lm: Training multi-billion parameter language models using model parallelism},
  author={Shoeybi, Mohammad and Patwary, Mostofa and Puri, Raul and LeGresley, Patrick and Casper, Jared and Catanzaro, Bryan},
  journal={arXiv preprint arXiv:1909.08053},
  year={2019}
}

@article{kingma2014adam,
  title={Adam: A method for stochastic optimization},
  author={Kingma, Diederik P and Ba, Jimmy},
  journal={arXiv preprint arXiv:1412.6980},
  year={2014}
}

@article{ziyin2020laprop,
  title={LaProp: Separating momentum and adaptivity in Adam},
  author={Ziyin, Liu and Wang, Zhikang T and Ueda, Masahito},
  journal={arXiv preprint arXiv:2002.04839},
  year={2020}
}

@article{hinton2012neural,
  title={Neural networks for machine learning lecture 6a overview of mini-batch gradient descent},
  author={Hinton, Geoffrey and Srivastava, Nitish and Swersky, Kevin},
  journal={Cited on},
  volume={14},
  number={8},
  pages={2},
  year={2012}
}

@article{zhao2026towards,
  title={Towards a Principled Muon under $\mu$P: Ensuring Spectral Conditions throughout Training},
  author={Zhao, John},
  journal={arXiv preprint arXiv:2601.01306},
  year={2026}
}

@article{du2026newton,
  title={The Newton-Muon Optimizer},
  author={Du, Zhehang and Su, Weijie},
  journal={arXiv preprint arXiv:2604.01472},
  year={2026}
}

@article{zhang2026mousse,
  title={Mousse: Rectifying the Geometry of Muon with Curvature-Aware Preconditioning},
  author={Zhang, Yechen and Xing, Shuhao and Huang, Junhao and Lv, Kai and Zhou, Yunhua and Qiu, Xipeng and Guo, Qipeng and Chen, Kai},
  journal={arXiv preprint arXiv:2603.09697},
  year={2026}
}

@article{zhang2026muon+,
  title={Muon+: Towards Better Muon via One Additional Normalization Step},
  author={Zhang, Ruijie and Zhao, Yequan and Liu, Ziyue and Wang, Zhengyang and Zhang, Zheng},
  journal={arXiv preprint arXiv:2602.21545},
  year={2026}
}

@article{zhang2026adam,
  title={Adam Improves Muon: Adaptive Moment Estimation with Orthogonalized Momentum},
  author={Zhang, Minxin and Liu, Yuxuan and Schaeffer, Hayden},
  journal={arXiv preprint arXiv:2602.17080},
  year={2026}
}

@article{yang2026manifold,
  title={Manifold constrained steepest descent},
  author={Yang, Kaiwei and Lai, Lexiao},
  journal={arXiv preprint arXiv:2601.21487},
  year={2026}
}

@article{everett2024scaling,
  title={Scaling exponents across parameterizations and optimizers},
  author={Everett, Katie and Xiao, Lechao and Wortsman, Mitchell and Alemi, Alexander A and Novak, Roman and Liu, Peter J and Gur, Izzeddin and Sohl-Dickstein, Jascha and Kaelbling, Leslie Pack and Lee, Jaehoon and others},
  journal={arXiv preprint arXiv:2407.05872},
  year={2024}
}

@article{wortsman2023small,
  title={Small-scale proxies for large-scale transformer training instabilities},
  author={Wortsman, Mitchell and Liu, Peter J and Xiao, Lechao and Everett, Katie and Alemi, Alex and Adlam, Ben and Co-Reyes, John D and Gur, Izzeddin and Kumar, Abhishek and Novak, Roman and others},
  journal={arXiv preprint arXiv:2309.14322},
  year={2023}
}

@article{yang2023tensor,
  title={Tensor programs ivb: Adaptive optimization in the infinite-width limit},
  author={Yang, Greg and Littwin, Etai},
  journal={arXiv preprint arXiv:2308.01814},
  year={2023}
}

@article{magakyan2026spectral,
  title={Spectral Scaling Laws of Muon},
  author={Magakyan, Gagik and Parrilo, Pablo and Ozdaglar, Asuman},
  journal={arXiv preprint arXiv:2606.04058},
  year={2026}
}

@online{epsilon_nuisance,
  author = {Nado, Zachary},
  title  = {{\(\epsilon\)}, a Nuisance no more. \texttt{https://zna.do/epsilon}},
}

@inproceedings{sutskever2013importance,
  title={On the importance of initialization and momentum in deep learning},
  author={Sutskever, Ilya and Martens, James and Dahl, George and Hinton, Geoffrey},
  booktitle={International conference on machine learning},
  pages={1139--1147},
  year={2013},
  organization={pmlr}
}

@article{dozat2016incorporating,
  title={Incorporating nesterov momentum into adam},
  author={Dozat, Timothy},
  booktitle={OpenReview},
  year={2016}
}

%%%%%%%%%%%%%%%%%%%%%%%%%%%%%%%%%%%%%%%%%%%%%%%%%%%%%%%%%%%%
\newpage
\appendix

\section{Appendix}
\subsection{An SVD perspective on SOAP's preconditioner: Relating Muon, SOAP, and Shampoo}\label{svd_muon_soap}

As has been studied before, singular value decomposition can give us insight into how Shampoo, SOAP, and Muon are related to each other \cite{shah2025practical}.

Let us examine the SVD of the averaged weight gradient: \footnote{An ill-conditioned gradient will have singular values with a large $\max/\min$ ratio: $\dfrac{\max(\Sigma)}{\min(\Sigma)} >> 1$}
\begin{equation}
\mathbf{G}_t = \mathbf{U} \cdot \mathbf{\Sigma} \cdot \mathbf{V}^T.
\end{equation}

Now notice (with EMAs turned off for simplicity):

\begin{itemize}
    \item $\mathbf{Q}_L$ is the eigenvector matrix of $\mathbf{G}_t\,\mathbf{G}_t^T$, which is $\mathbf{U}$, since
    \begin{equation}\label{eq_left_svd}
    \mathbf{G}_t\,\mathbf{G}_t^T = \mathbf{U}\,\mathbf{\Sigma}^2\,\mathbf{U}^T.
    \end{equation}
    \item $\mathbf{Q}_R$ is the eigenvector matrix of $\mathbf{G}_t^T\,\mathbf{G}_t$, which is $\mathbf{V}$, since
    \begin{equation}\label{eq_right_svd}
    \mathbf{G}_t^T\,\mathbf{G}_t = \mathbf{V}\,\mathbf{\Sigma}^2\,\mathbf{V}^T.
     \end{equation}
\end{itemize}

We have
$$
\mathbf{G}^{R}_t = \mathbf{Q}_L^T\,\mathbf{G}_t\,\mathbf{Q}_R = 
\underbrace{\left(\mathbf{U}^T\,\mathbf{U}\right)}_{=\,\mathbf{I}_{\text{in}\times \text{in}}}
\,\mathbf{\Sigma}\,
\underbrace{\left(\mathbf{V}^T\,\mathbf{V}\right)}_{=\,\mathbf{I}_{\text{out}\times \text{out}}} = \mathbf{\Sigma}
$$
Thus the gradient rotated to the eigenbasis of Shampoo's preconditioner effectively diagonalizes it. 
Now applying Adam to $\mathbf{G}^R_t$:
$$
\mathbf{N} = \operatorname{Adam}(\mathbf{G}'_t) = \operatorname{Adam}(\mathbf{\Sigma}).
$$

The weight update is then given by:
$$
\Delta \mathbf{W} = -\eta\, \mathbf{Q}_L\,\mathbf{N}\,\mathbf{Q}_R^T = -\eta\, \mathbf{U}\,\mathbf{N}\,\mathbf{V}^T = -\eta\, \mathbf{U}\cdot \operatorname{Adam}(\mathbf{\Sigma})\cdot \mathbf{V}^T.
$$

Considering the limit where Adam can be approximated by the \texttt{sign} function \cite{balles2018dissecting} and when EMA is turned off, the eigenvector matrices perfectly diagonalize the gradient, $\texttt{sign}(\Sigma) = \text{Id}_{\text{min}(\text{in},\text{out}) \times \text{min}(\text{in},\text{out})}$, the update is given by the polar factor:
\begin{equation}\label{soap_whiten}
    \Delta \mathbf{W} = -\eta\, \mathbf{U}\cdot \operatorname{Adam}(\mathbf{\Sigma})\cdot \mathbf{V}^T \approx -\eta\, \mathbf{U}\cdot \texttt{sign}(\mathbf{\Sigma})\cdot \mathbf{V}^T = -\eta\, \mathbf{U} \, \mathbf{V}^T
\end{equation}

Similarly, consider Shampoo with exponent $p= -1/4$ \cite{gupta2018shampoo}:
\begin{equation*}
\Delta \mathbf{W} = -\eta \mathbf{(GG^T)}^{-1/4} \, \mathbf{G} \, \mathbf{(G^T G)}^{-1/4}.
\end{equation*}

Taking the $-\frac{1}{4}$ power of each term from Eq.~\ref{eq_left_svd} and Eq.~\ref{eq_right_svd}, we obtain
\begin{align*}
\mathbf{(GG^T)}^{-1/4} &= \bigl(\mathbf{U} \, \mathbf{\Sigma}^2 \, \mathbf{U}^T\bigr)^{-1/4} = \mathbf{U} \, \mathbf{\Sigma}^{-1/2} \, \mathbf{U}^T, \\
\mathbf{(G^T G)}^{-1/4} &= \bigl(\mathbf{V} \, \mathbf{\Sigma}^2 \, \mathbf{V}^T\bigr)^{-1/4} = \mathbf{V} \, \mathbf{\Sigma}^{-1/2} \, \mathbf{V}^T.
\end{align*}

Thus, substituting into the update expression for Shampoo,
\begin{align*}
\mathbf{(GG^T)}^{-1/4} \, \mathbf{G} \, \mathbf{(G^T G)}^{-1/4}
&= \Bigl(\mathbf{U} \, \mathbf{\Sigma}^{-1/2} \, \mathbf{U}^T\Bigr)
\Bigl(\mathbf{U} \, \mathbf{\Sigma} \, \mathbf{V}^T\Bigr)
\Bigl(\mathbf{V} \, \mathbf{\Sigma}^{-1/2} \, \mathbf{V}^T\Bigr) \\
&= \mathbf{U} \, \mathbf{\Sigma}^{-1/2} \, \underbrace{\mathbf{U}^T \mathbf{U}}_{\mathbf{I}_{\text{in}\times \text{in}}} \, \mathbf{\Sigma} \, \underbrace{\mathbf{V}^T \mathbf{V}}_{\mathbf{I}_{\text{out}\times \text{out}}} \, \mathbf{\Sigma}^{-1/2} \, \mathbf{V}^T \\
&= \mathbf{U} \, (\mathbf{\Sigma}^{-1/2} \, \mathbf{\Sigma} \, \mathbf{\Sigma}^{-1/2} \,) \mathbf{V}^T \\
&= \mathbf{U} \, (\mathbf{Id}) \, \mathbf{V}^T \\
&= \mathbf{U} \, \mathbf{V}^T.
\end{align*}

Hence, Shampoo's preconditioner is whitening the gradient, which matches what SOAP and Muon are doing in the simplified SVD view without momentum, eq.~\ref{soap_whiten}:
$$
\Delta \mathbf{W} = -\eta \mathbf{(GG^T)}^{-1/4} \, \mathbf{G} \, \mathbf{(G^T G)}^{-1/4} = -\eta \mathbf{U} \, \mathbf{V}^T.
$$

\subsection{An SVD Perspective on Shampoo's Hessian: Whitening Curvature}

For a weight parameter $W \in \mathcal{R}^{\text{in} \times \text{out}}$, the optimal Kronecker approximation to the Gauss-Newton Hessian \cite{morwani2024new} was shown to be related to Shampoo's preconditioner:
\begin{equation*}\label{shampoo_hessian}
H = \frac{\mathbb{E}\left[G\,G^\top\right] \otimes \mathbb{E}\left[G^\top\,G\right]}{\operatorname{Tr}\left(\mathbb{E}\left[G\,G^\top\right]\right)}
\end{equation*}

Substituting the spectral decompositions of the covariance matrices from Eqs.~\ref{eq_left_svd} and \ref{eq_right_svd} into the Kronecker approximation (assuming the expectations are approximated by the current statistics), we obtain:

\begin{equation}
\mathbf{H} \propto (\mathbf{U} \mathbf{\Sigma}^2 \mathbf{U}^T) \otimes (\mathbf{V} \mathbf{\Sigma}^2 \mathbf{V}^T).
\end{equation}

Using the mixed-product property of the Kronecker product, $(\mathbf{A} \otimes \mathbf{B})(\mathbf{C} \otimes \mathbf{D}) = (\mathbf{AC}) \otimes (\mathbf{BD})$, we can factorize the expression into an eigendecomposition of the Hessian:

\begin{equation} \mathbf{H} \propto \underbrace{(\mathbf{U} \otimes \mathbf{V})}_{\text{Eigenvectors}} \cdot \underbrace{(\mathbf{\Sigma}^2 \otimes \mathbf{\Sigma}^2)}_{\text{Eigenvalues}} \cdot \underbrace{(\mathbf{U}^\top \otimes \mathbf{V}^\top)}_{\text{Basis Change}}. \end{equation}

This reveals the spectral structure of the implicit Hessian used by Shampoo and SOAP:
\begin{itemize}
    \item The \textbf{eigenvectors} of Shampoo's approximate Hessian are the Kronecker products of the left and right singular vectors of the gradient, $\mathbf{u}_i \otimes \mathbf{v}_j$.
    \item The \textbf{eigenvalues} (curvature) are the products of the squared singular values, $\lambda_{ij} \propto \sigma_i^2 \sigma_j^2$.
\end{itemize}

Consequently, the preconditioning operation (which effectively applies $\mathbf{H}^{-1/4}$ in Shampoo or $\mathbf{H}^{-1/2}$ in SOAP) dampens the gradient updates primarily along directions corresponding to large singular values (where curvature $\sigma_i^2 \sigma_j^2$ is high) and amplifies them where singular values are small.

\subsection{An SVD perspective on KL-Shampoo: Improving Conditioning for Inversion and Decomposition}\label{appendex:kl_condition}

While Shampoo and SOAP update Kronecker factors using the marginal covariance statistics $\mathbf{G}\mathbf{G}^T$ and $\mathbf{G}^T\mathbf{G}$, KL-Shampoo \cite{lin2025understanding} introduces a coupled update rule where each factor is updated using the inverse of the other. Here we analyze the spectral properties of this update using the singular value decomposition (SVD) of the gradient.

Let the SVD of the gradient at step $t$ be $\mathbf{G}_t = \mathbf{U} \mathbf{\Sigma} \mathbf{V}^T$. We assume the Kronecker factors $\mathbf{S}_a$ and $\mathbf{S}_b$ are approximately aligned with the gradient's eigenbasis, such that:
\begin{align*}
\mathbf{S}_a &\approx \mathbf{U} \mathbf{\Lambda}_a \mathbf{U}^T \\
\mathbf{S}_b &\approx \mathbf{V} \mathbf{\Lambda}_b \mathbf{V}^T
\end{align*}
where $\mathbf{\Lambda}_a$ and $\mathbf{\Lambda}_b$ represent the eigenvalues captured by the Kronecker factors.

Recall the KL-Shampoo update rule for $\mathbf{S}_a$:
\begin{equation}
\mathbf{S}_a \leftarrow (1-\beta)\mathbf{S}_a + \beta \left( \mathbf{G} \mathbf{S}_b^{-1} \mathbf{G}^T \right).
\end{equation}

Substituting the spectral forms into the update term:
\begin{align*}
\mathbf{G} \mathbf{S}_b^{-1} \mathbf{G}^T &= (\mathbf{U} \mathbf{\Sigma} \mathbf{V}^T) (\mathbf{V} \mathbf{\Lambda}_b^{-1} \mathbf{V}^T) (\mathbf{V} \mathbf{\Sigma} \mathbf{U}^T) \\
&= \mathbf{U} \mathbf{\Sigma} (\mathbf{V}^T \mathbf{V}) \mathbf{\Lambda}_b^{-1} (\mathbf{V}^T \mathbf{V}) \mathbf{\Sigma} \mathbf{U}^T \\
&= \mathbf{U} (\mathbf{\Sigma} \mathbf{\Lambda}_b^{-1} \mathbf{\Sigma}) \mathbf{U}^T \\
&= \mathbf{U} (\mathbf{\Sigma}^2 \mathbf{\Lambda}_b^{-1}) \mathbf{U}^T.
\end{align*}

At a fixed point where the preconditioning statistics stabilize ($\mathbf{S}_{new} \approx \mathbf{S}_{old}$), the eigenvalues of the updated factor must match the eigenvalues of the update term. Thus, for the diagonal entries:
\begin{equation}
\mathbf{\Lambda}_a \approx \mathbf{\Sigma}^2 \mathbf{\Lambda}_b^{-1}.
\end{equation}

By symmetry, the update for $\mathbf{S}_b$ yields $\mathbf{\Lambda}_b \approx \mathbf{\Sigma}^2 \mathbf{\Lambda}_a^{-1}$. Solving this system for the case where the spectra are balanced ($\mathbf{\Lambda}_a \approx \mathbf{\Lambda}_b = \mathbf{\Lambda}$):
\begin{align*}
\mathbf{\Lambda} &\approx \mathbf{\Sigma}^2 \mathbf{\Lambda}^{-1} \\
\mathbf{\Lambda}^2 &\approx \mathbf{\Sigma}^2 \\
\mathbf{\Lambda} &\approx \mathbf{\Sigma}.
\end{align*}

This result highlights the fundamental difference between standard Shampoo and KL-Shampoo:
\begin{itemize}
    \item \textbf{Shampoo:} Kronecker factors track the covariance, so $\mathbf{\Lambda}_{\text{Shampoo}} \approx \mathbf{\Sigma}^2$.
    \item \textbf{KL-Shampoo:} Kronecker factors track the singular values, so $\mathbf{\Lambda}_{\text{KL-Shampoo}} \approx \mathbf{\Sigma}$.
\end{itemize}

This has an important advantage for numerical stability. The condition number of the matrices being inverted (or decomposed) in KL-Shampoo is the square root of that in standard Shampoo:
\begin{equation}
\kappa(\mathbf{S}_{\text{KL-Shampoo}}) = \frac{\sigma_{\max}}{\sigma_{\min}} = \sqrt{\kappa(\mathbf{S}_{\text{Shampoo}})}.
\end{equation}
For ill-conditioned gradients where $\kappa(\mathbf{G})$ is large, KL-Shampoo operates on significantly better-conditioned matrices, reducing the accumulation of numerical errors during the eigen-decomposition steps.

\section{MXFP8 Pretraining}\label{app_sec:mxfp8}

We use the MXFP8 recipe developed by NVIDIA for Blackwell \cite{mishra2025recipes} together with Muon and pretrain NVIDIA-Nemotron-NanoV3 and the 8B Hybrid Mamba-Attention LatentMoE for 1T tokens. We find that MXFP8 has a small loss gap, while most downstream evaluations are higher with MXFP8 than BF16, Table~\ref{tab:precision_ablation_results}. Further, we find that replacing AdamW with Lion \cite{chen2023symbolic} with $(\beta_1, \beta_2)= (0.95, 0.98)$ improves losses and evaluations (Table~\ref{tab:precision_ablation_results}, last column). 

\begin{table}
\centering
\resizebox{0.75\textwidth}{!}{%
\begin{tabular}{lccccc}
\toprule
\textbf{Metric} 
& \multicolumn{2}{c}{\textbf{NanoV3-1T}} 
& \multicolumn{3}{c}{\textbf{8B Hybrid LatentMoE-1T}} \\
\cmidrule(lr){2-3} \cmidrule(lr){4-6}
& BF16 & MXFP8 & BF16 & MXFP8 & MXFP8 (Lion) \\
\midrule

\multicolumn{6}{c}{\textbf{MMLU}} \\
\midrule
MMLU & \textbf{69.33} & 67.84 & 74.89 & 75.7 & \textbf{76.00} \\
MMLU PRO CoT & \textbf{51.20} & 50.37 & 59.6 & 61.6 & \textbf{62.2} \\

\midrule
\multicolumn{6}{c}{\textbf{Coding}} \\
\midrule
HumanEval & 57.68 & \textbf{59.12} & \textbf{65.79} & 64.8 & 65.5 \\
HumanEval+ & 52.38 & \textbf{55.61} & \textbf{61.86} & 61.2 & 61.5 \\
MBPP & 63.72 & \textbf{64.12} & 70.99 & 71.3 & \textbf{72.4} \\
MBPP+ & 52.41 & \textbf{56.26} & 55.26 & 58.77 & \textbf{60.6} \\
\cellcolor{gray!15}Coding Avg. & \cellcolor{gray!15}56.55 & \cellcolor{gray!15}\textbf{58.78} & \cellcolor{gray!15}63.48 & \cellcolor{gray!15}64.02 & \cellcolor{gray!15}\textbf{65.00} \\

\midrule
\multicolumn{6}{c}{\textbf{Math}} \\
\midrule
Math 500 & \textbf{69.45} & 67.25 & 75.6 & 77.1 & \textbf{78.75} \\
GSM8k CoT & \textbf{85.82} & 85.37 & 85.97 & \textbf{89.2} & 86.5 \\
\cellcolor{gray!15}Math Avg. & \cellcolor{gray!15}\textbf{77.64} & \cellcolor{gray!15}76.31 & \cellcolor{gray!15}80.79 & \cellcolor{gray!15}\textbf{83.15} & \cellcolor{gray!15}82.63 \\

\midrule
\multicolumn{6}{c}{\textbf{Commonsense Reasoning}} \\
\midrule
Race & 82.49 & 82.49 & 86.32 & 87.1 & \textbf{88.6} \\
ARC Challenge & 83.10 & \textbf{84.64} & 89.16 & 89.4 & \textbf{92.00} \\
Winogrande & 70.09 & \textbf{71.35} & \textbf{75.37} & 75.1 & 74.8 \\
Hellaswag & \textbf{79.36} & 79.34 & 82.37 & 82.5 & \textbf{83.2} \\
\cellcolor{gray!15}Commonsense Avg. & \cellcolor{gray!15}78.76 & \cellcolor{gray!15}\textbf{79.46} & \cellcolor{gray!15}83.31 & \cellcolor{gray!15}83.53 & \cellcolor{gray!15}\textbf{84.65} \\

\bottomrule
\end{tabular}%
}
\caption{Benchmark results across MMLU, Coding, Math, and Commonsense Reasoning comparing BF16 and MXFP8 precision formats for the NVIDIA-Nemotron-NanoV3-1T and 8B Hybrid LatentMoE-1T models. All experiments were performed at $2\times$GBS of 25M tokens with Muon as the backbone optimizer.}
\label{tab:precision_ablation_results}
\end{table}

\begin{figure}
    \centering
    \includegraphics[width=\textwidth, height=0.85\textheight, keepaspectratio]{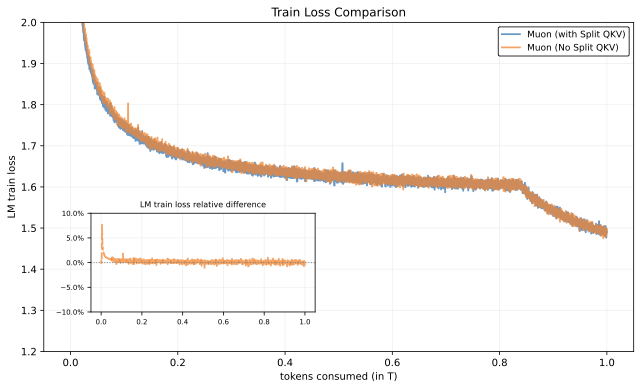}
    \caption{Training loss of a Qwen-3-30B-A3B architecture comparing Muon's Newton-Schulz with and without splitting the fused QKV tensor. With split-QKV, the model has lower training loss early in pretraining, but this difference disappears as we scale token counts.}
    \label{fig:nanov3-splitqkv}
\end{figure}

\section{AdamW at Larger Batch Sizes}
\begin{figure}
    \centering
    \includegraphics[width=0.75\linewidth]{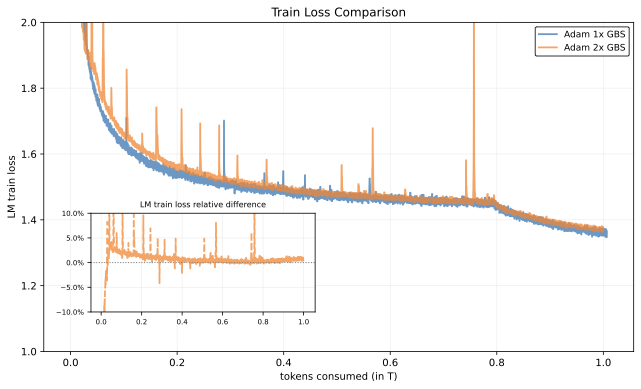}
    \caption{NVIDIA-Nemotron-NanoV3 pretrained with AdamW for a 1T-token horizon at 25M and 50M batch sizes. Batch size ramp-up and square-root scaling were applied to the 50M batch size run.}
    \label{fig:sup_adamW_2xgbs}
\end{figure}

We find that the techniques that enabled larger batch pretraining with Muon did not help with AdamW, Fig.~\ref{fig:sup_adamW_2xgbs}. We leave tuning the hyperparameters $\beta_1, \beta_2, \epsilon$ of AdamW with batch size as future work. 

\section{Optimizing Conv1d Filters with Muon}

\begin{figure}
    \centering
    \includegraphics[width=\textwidth, height=0.85\textheight, keepaspectratio]{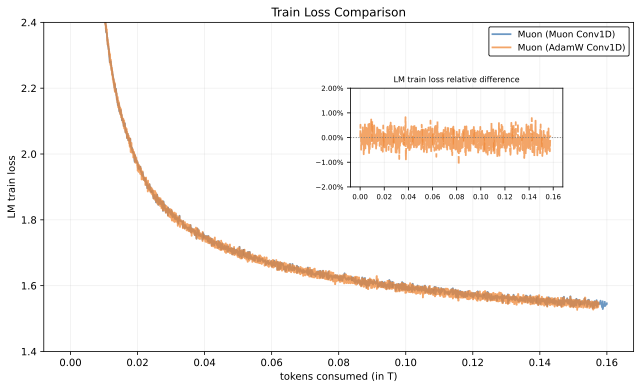}
    \caption{Training loss of Nano-V3 comparing the application of Muon vs. AdamW on the Mamba mixer Conv1D. Using AdamW on the Conv1D layer leads to slightly lower ($\approx 0.1\%$) loss.}
    \label{fig:nanov3-adamw-conv1d}
\end{figure}

For hybrid architectures incorporating Mamba2 SSM layers, we found that excluding \texttt{Conv1D}c parameters from Muon and optimizing them with AdamW instead yielded a modest but consistent improvement in training and validation loss, as shown in Fig.~\ref{fig:nanov3-adamw-conv1d}. The Conv1D weights in Mamba2 encode localized temporal filters that are shared and re-applied at every sequence position, making orthogonality constraints geometrically ill-motivated compared to standard linear projections. 

\subsection{Splitting fused QKV tensors before orthogonalization}
Fig.~\ref{fig:nanov3-splitqkv} shows Muon with and without splitting QKV tensors before orthogonalization. While splitting achieves lower losses early in training, the losses at the end of the full 1T-token horizon are identical. 

\section{SOAP Eigenbasis Update Frequency}

\begin{figure}
    \centering
    \includegraphics[width=\textwidth, height=0.85\textheight, keepaspectratio]{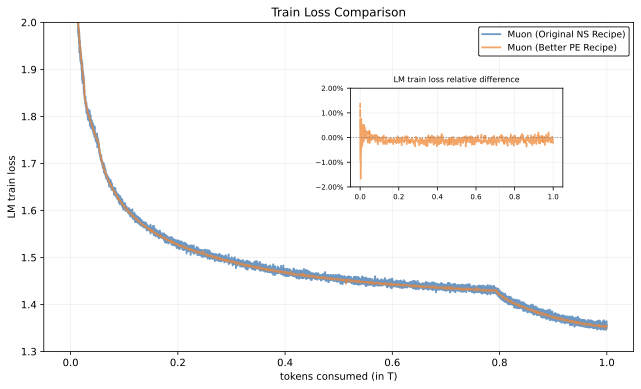}
    \caption{Training loss of Nano-V3 comparing Muon's quintic-coefficient optimized Newton-Schulz recipe and the Polar Express recipe run for 16 iterations.}
    \label{fig:nanov3-polarexpress-16}
\end{figure}

\begin{figure}
    \centering
    \includegraphics[width=\textwidth, height=0.85\textheight, keepaspectratio]{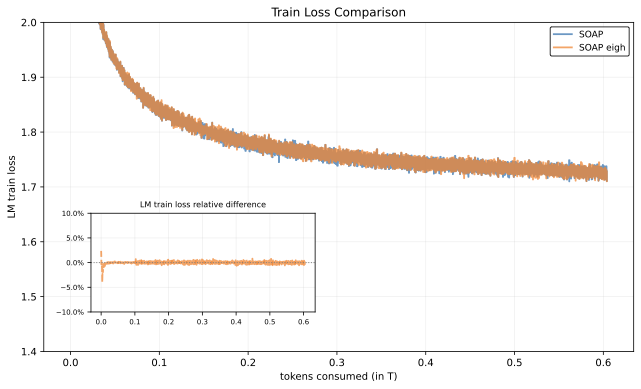}
    \caption{Using \texttt{eigh} vs. \texttt{QR} decomposition to recompute SOAP's eigenbasis has a negligible impact on large-scale pretraining loss on a Qwen-3-30B-A3B architecture; hence, we default to using \texttt{QR} since it is cheaper.}
    \label{fig:eigh_vs_QR_soap}
\end{figure}

Fig.~\ref{fig:eigh_vs_QR_soap} is a comparison of \texttt{eigh} vs. \texttt{QR} decomposition to recompute the eigenbasis in SOAP. We found it necessary to perform QR decomposition at every step to reduce the loss spikes. This is likely due to the rapidly changing nature of the preconditioner basis early in training, which slower update schedules failed to capture. With this frequency, there was negligible difference between the use of \texttt{QR} or \texttt{eigh}. 

\end{document}